\documentclass[conference]{IEEEtran}
\IEEEoverridecommandlockouts

\usepackage{cite}
\usepackage{flushend}
\usepackage{caption}
\usepackage{tabularx}
\usepackage{graphicx}
\usepackage{subfigure}
\usepackage{amsfonts}
\usepackage{amsmath}
\usepackage{amsthm}
\usepackage{algorithm}
\usepackage{algorithmic}
\usepackage{multirow}
\usepackage{xcolor}
\usepackage{booktabs}
\usepackage{float}
\usepackage{bm}
\usepackage{xspace}
\usepackage{threeparttable}
\usepackage{amssymb}
\usepackage{url}
\usepackage[misc]{ifsym}
\usepackage{enumerate}
\usepackage{balance}

\usepackage[hidelinks]{hyperref}

\newcommand{\ie}{\textit{i}.\textit{e}.}
\newcommand{\eg}{\textit{e}.\textit{g}.} 
\newcommand{\wrt}{\textit{w}.\textit{r}.\textit{t}} 
\newtheorem{Def}{Definition} 
\newtheorem{Problem}{Problem}

\def\framework{HARE\xspace}
\def\model{FAST\xspace}
\def\models{\model-Star\xspace}
\def\modelt{\model-Tri\xspace}
\def\modelp{\model-Pair\xspace}

\def\frameworkp{HARE-Pair\xspace}

\ifCLASSINFOpdf
\else
\fi
\def\BibTeX{{\rm B\kern-.05em{\sc i\kern-.025em b}\kern-.08em
    T\kern-.1667em\lower.7ex\hbox{E}\kern-.125emX}}

\begin{document}


\title{Scalable Motif Counting for Large-scale \\ Temporal Graphs}

\author{
\IEEEauthorblockN{
Zhongqiang Gao{\textsuperscript{1}*}, Chuanqi Cheng{\textsuperscript{1}*},  Yanwei Yu{\textsuperscript{1, \Letter}}, Lei Cao{\textsuperscript{2}}, Chao Huang{\textsuperscript{3}},  Junyu Dong{\textsuperscript{1}}}\\
\IEEEauthorblockA{
\textsuperscript{1}College of Computer Science and Technology, Ocean University of China, Qingdao, China\\
\textsuperscript{2}Computer Science and Artificial Intelligence Laboratory, Massachusetts Institute of Technology, Cambridge, USA\\
\textsuperscript{3}Department of Computer Science, The University of Hong Kong, Hong Kong, China\\
\{gaozhongqiang, chengchuanqi\}@stu.ouc.edu.cn, \{yuyanwei, dongjunyu\}@ouc.edu.cn, \\lcao@csail.mit.edu, chuang@cs.hku.hk
}\\
\thanks{*Authors contribute this work equally. Corresponding author: Yanwei Yu.}
}


\maketitle

\begin{abstract}

One fundamental problem in temporal graph analysis is to count the occurrences of small connected subgraph patterns (\ie, motifs), which benefits a broad range of real-world applications, such as anomaly detection, structure prediction, and network representation learning. However, existing works focused on exacting temporal motif are not scalable to large-scale temporal graph data, due to their heavy computational costs or inherent inadequacy of parallelism. In this work, we propose a scalable parallel framework for exactly counting temporal motifs in large-scale temporal graphs. We first categorize the temporal motifs based on their distinct properties, and then design customized algorithms that offer efficient strategies to exactly count the motif instances of each category. 
Moreover, our compact data structures, namely triple and quadruple counters, enable our algorithms to directly identify the temporal motif instances of each category, according to edge information and relationship between edges, therefore significantly improving the counting efficiency. 
Based on the proposed counting algorithms, we design a hierarchical parallel framework that featuring both inter- and intra-node parallel strategies, and fully leverages the multi-threading capacity of modern CPU to concurrently count all temporal motifs. 
Extensive experiments on sixteen real-world temporal graph datasets demonstrate the superiority and capability of our proposed framework for temporal motif counting, achieving up to $538\times$ speedup compared to the state-of-the-art methods. 
The source code of our method is available at:~\url{https://github.com/steven-ccq/FAST-temporal-motif}.  
\end{abstract}



\section{Introduction}  
\label{section:introduction}

Many real-world applications are naturally represented in a graph data structure such as social networks, traffic networks, citation networks, biology networks, and knowledge graphs, where objects and the relationships among them are respectively represented by nodes and edges. 
In real-world scenarios, many networks constantly evolve over time with their structures dynamically evolving as new relationships constantly emerges. Such dynamic networks are termed as \textit{temporal graphs}~\cite{paranjape2017motifs,liu2021temporal} composed of a set of nodes and a series of timestamped edges between nodes, or temporal edges. Examples include email networks, communication networks, financial transactions, and E-commercial networks.

Counting patterns in graph data is one of fundamental problems in graph data mining, widely used in a variety of network analytical tasks such as anomaly detection~\cite{akoglu2015graph}, role discovery~\cite{ahmed2020role}, and community detection~\cite{fang2020survey}. 
An especially useful case is counting \textit{motifs} (or \textit{graphlets}) -- a category of frequent subgraph patterns, which are used in range of disciplines, including social network analysis~\cite{yang2018node}, neuroscience~\cite{hu2013motif} and computational biology~\cite{prvzulj2007biological}. 
For example, social network analysis often uses communication motifs mined from large dynamic networks to understand how human communication unfold~\cite{gurukar2015commit}. 
Moreover, because motif is effective in capturing local high-order network structures, 
recently leveraging motif to improve the quality of network embedding has attracted great attention~\cite{liu2021motif,huang2020motif,jin2020gralsp,yu2019rum}.  


In this work, we target on designing a {\it scalable}, parallel solution to efficiently count temporal motifs from large-scale dynamic graphs. 
The existing works that attempt to count temporal motifs provide either {\it exact} results or {\it approximations}. 
The existing exact algorithms can hardly handle large-scale network, due to their heavy computation costs. 
Paranjape~\textit{et al.}~\cite{paranjape2017motifs} formally define the notion of \textit{$\delta$-temporal motifs}, and propose an exact algorithm (EX) for counting 2- and 3-node, 3-edge, $\delta$-temporal motifs by leveraging subgraph enumeration. Kumar and Calders~\cite{kumar20182scent} present an efficient algorithm called 2SCENT to find all temporal cycles in a directed interaction network. 
Mackey~\textit{et al.}~\cite{mackey2018chronological} propose an efficient backtracking (BT) algorithm for temporal subgraph isomorphism, which can exactly count temporal motifs by enumerating all of them. 
However, because they have to enumerate all edges and circles, these works still suffer from high computation complexity, hence not scalable to big graph.
For example, when counting temporal motifs on the RedditComments data with more than 600 million edges, it takes EX 7,968 seconds to find all 2- and 3-node, 3-edge, $\delta$-temporal motifs.
This high response time makes such methods insufficient in handling frequently updated dynamic systems which are very popular in practice.  

To reduce response time, several sampling-based algorithms have been proposed to {\it approximate} the number of motifs~\cite{liu2019sampling,wang2020efficient}. 
However, these approximate algorithms, either only supporting certain types of motifs such as 2-node 3-edge motifs or suffering from large approximate errors, do not meet the requirements of many applications. 
Furthermore, the motifs discovered by sampling fail to preserve the local structures of a graph~\cite{yu2019rum}. Therefore, they are not effective when used in network embedding, which is one of the most important emerging applications of motifs.  



To address the aforementioned challenges, we propose a scalable parallel framework called  \framework for temporal motif counting in large-scale temporal networks. 
First, based on topological structure, we categorize all possible 2- and 3-node, 3-edge, $\delta$-temporal motifs into three types: \textit{pair temporal motifs}, \textit{star temporal motifs} and \textit{triangle temporal motifs}. 
Customized to star/pair temporal motifs, our \models algorithm uses a quadruple counter and a triple counter to compactly encode the number of star motif instances and pair motif instances, respectively. 
With the designed counters, \models directly identifies the types of temporal motif instances according to the information of edges and the relationship between edges, therefore, significantly improving the counting efficiency.
Our algorithm \modelt, customized to triangle temporal motifs, uses another quadruple counter to record the number of motif instances for the non-isomorphic temporal motifs simultaneously.  
\model (general term for \models and \modelt) recursively treats each node in the given temporal graph as center node, and searches all motif instances in the edge sequence of the center node, achieving a linear time complexity in the number of temporal edges of input graph. 
Furthermore, the recursive nature of \model enables us to leverage multi-threading of modern CPU to count the temporal motifs in parallel. This is because in our parallel framework \framework, if different threads pick different centers, each thread will exactly count distinct motifs independently. 



We conduct extensive experiments on 16 real-world large-scale temporal graphs, and the experimental results demonstrate that our \framework performs significantly faster than state-of-the-art baselines for counting temporal motifs by up to two orders of magnitudes. 

We highlight the key contributions of this work as follows:

\begin{itemize}
    \item We propose a fast exact algorithm for counting star/pair temporal motifs. The proposed \models algorithm can directly identify the types of motif instances according to the edge information and relationship between edges, improving the significant detection efficiency. 
    
    \item We develop a fast algorithm for exactly counting motif instances for triangle temporal motifs. The proposed \modelt algorithm can simultaneously count the number of motif instances for all non-isomorphic triangle temporal motifs with the designed quadruple counter. 
    
    \item We propose a hierarchical parallel framework HARE for our proposed two exact algorithms, which endows our method with the capability of concurrently counting all temporal motifs for large-scale temporal networks in an efficient way. 
    
    \item We perform extensive experiments on 16 real-world graph datasets to demonstrate the superiority of our proposed method compared with other baselines. Our proposed \framework results in up to two orders of magnitude faster than state-of-the-art techniques. 
\end{itemize}

\section{Related Work}

\subsection{Motif Counting in Static Graphs}
There have been rich studies on network motifs in static graphs, where these works have proved crucial to understanding the mechanisms driving complex systems~\cite{milo2002network} and characterizing classes of static graphs~\cite{vazquez2004topological,yaverouglu2014revealing}. In addition, the motifs are very important for understanding the high-order organization model in the graph~\cite{benson2015tensor,benson2016higher}. 
In terms of algorithm, a variety of researches are only used to calculate triangles in undirected static graphs~\cite{2008Main}.    
Ahmed~\textit{et al.}~\cite{2015Efficient} propose a fast algorithm for counting motifs of ${3,4}$-node that leverages a number of combinatorial arguments.  It significantly improves the scalability of motif counting.  
Santoso~\textit{et al.}~\cite{santoso2020efficient} propose an exact algorithm for enumerating 4-node motifs, such as 4-cycles, 4-cliques and diamonds, by leveraging the most efficient algorithm for triangle enumeration. 
Since many graphs are not static as the links between nodes dynamically change over time~\cite{HOLME201297}, the above methods fail to capture the richness of the temporal information in the data. 

\subsection{Motif Counting in Temporal Graphs}
Recently, the temporal motif is no longer limited by the snapshot, but has been extended to the network motif with time information~\cite{zhao2010communication}.   
Kovanen~\textit{et al.}~\cite{kovanen2011temporal} first present the definition of temporal motif which is widely used in Wikipedia network. 
Gurukar~\textit{et al.}~\cite{gurukar2015commit} propose COMMIT based on subsequence mining to identify the temporal motifs in the communication network. 
In~\cite{paranjape2017motifs}, Paranjape~\textit{et al.} formally define the notion of \textit{$\delta$-temporal motifs}. They also propose exact fast algorithms for counting 2- and 3-node, 3-edge, $\delta$-temporal motifs by leveraging subgraph enumeration in temporal graphs. 
Kumar and Calders~\cite{kumar20182scent} focus on one such fundamental interaction pattern, namely a temporal cycle, and present an efficient algorithm called 2SCENT to find all temporal cycles in a directed interaction network. 
Mackey~\textit{et al.}~\cite{mackey2018chronological} propose an efficient backtracking algorithm for temporal subgraph isomorphism, which can count temporal motifs exactly by enumerating all of them.  
Based on the definition of communication motif in~\cite{gurukar2015commit}, Sun~\textit{et al.}~\cite{sun2019tm} propose an algorithm called TM-Miner, which can build a canonical labeling system that uses a new lexicographic order and maps the temporal graph to the unique minimum time first search code, to mining temporal motifs in large temporal network.

\subsection{Sampling Methods for Motif Counting}
First of all, many sampling methods have been proposed for approximate triangle counting in static graphs, such as subgraph sampling~\cite{tsourakakis2009doulion}, edge sampling~\cite{ahmed2014graph}, wedge sampling~\cite{turk2019revisiting} and neighborhood sampling~\cite{pavan2013counting}. 
Bera~\textit{et al.}~\cite{KDD20Triangle} propose a sublinear algorithm in the random walk access model to count triangles without seeing the whole static graph. Moreover, sampling methods are also efficient to find more complex motifs, \eg, 4-vertex motifs~\cite{sanei2018butterfly}, 5-vertex motifs~\cite{wang2017moss}, and $k$-cliques~\cite{jain2017fast}. However, all above methods do not consider the temporal information, and thus they can not process motif counting in temporal graphs directly. 
Recently, some sampling methods are proposed to approximately count motifs in temporal graphs. 
Liu~\textit{et al.}~\cite{liu2019sampling} develop a sampling framework that sits as a layer on top of existing exact counting algorithms.   
Wang~\textit{et al.}~\cite{wang2020efficient} propose an edge sampling algorithm for any temporal motifs and hybridize edge sampling with wedge sampling to count temporal motifs with 3 nodes and 3 edges.

\section{Problem Definition}

In this section, we first introduce key notations used in this work and then formally define the studied problem. 

\begin{Def}[Temporal Graph]
\label{def.temporal_graphs} 
A temporal graph is a graph $\mathcal{G}= \{\mathcal{V}, \mathcal{E}, \mathcal{T}\}$, where $\mathcal{V}$ is the collection of nodes, $\mathcal{E}$ is the collection of edges between the nodes, and $\mathcal{T}$ is the collection of timestamps. Each edge $e_{ij}^t$ is a timestamped directed edge from node $v_i$ to node $v_j$, denoted by $(v_i,v_j,t)$, where $v_i, v_j \in \mathcal{V}$ and $t \in \mathcal{T}$. We term each edge as a temporal edge. 
\end{Def} 

\begin{figure}[h]
\begin{center}
\vspace{-5mm}
\includegraphics[width=0.77\linewidth]{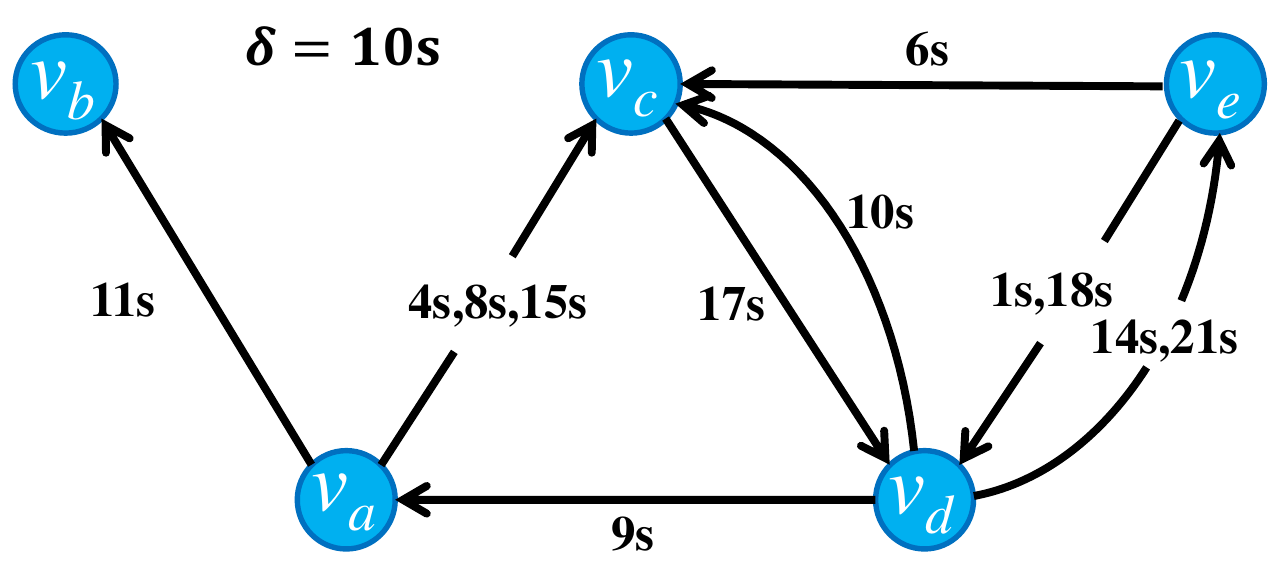}
\vspace{-2mm}
\caption{An example of a temporal graph.} 
\label{fig.toyexample}
\vspace{-4mm}
\end{center}
\end{figure}

Fig.~\ref{fig.toyexample} shows a toy example of a temporal graph with 5 nodes and 12 temporal edges, and each edge is directed and has a timestamp in seconds.


\begin{Def}[$\delta$-temporal Motif]
A $k$-node, $l$-edge, $\delta$-temporal motif is a sequence of $l$ temporal edges in chronological order within a $\delta$ time constraint, $M = \langle (u_1,v_1,t_1), (u_2,v_2,t_2), \dots, (u_l,v_l,t_l) \rangle$, \ie, $t_1\leq t_2 \leq \dots \leq t_l$ and $t_l-t_1 \leq \delta$, such that the induced static graph from these edges is connected and include $k$ nodes. 
\end{Def}


\begin{Def}[Motif Instance]
A collection of temporal edges selected from a given temporal graph is called a motif instance of a $\delta$-temporal motif $M$, if it matches the same edge pattern (\ie, same direction and time order) and all of the edges occur within $\delta$ time interval.  
\end{Def}

\begin{figure}[htp]
\begin{center}
\vspace{-3mm}
\hspace{-3mm}
\includegraphics[width=1.00\linewidth]{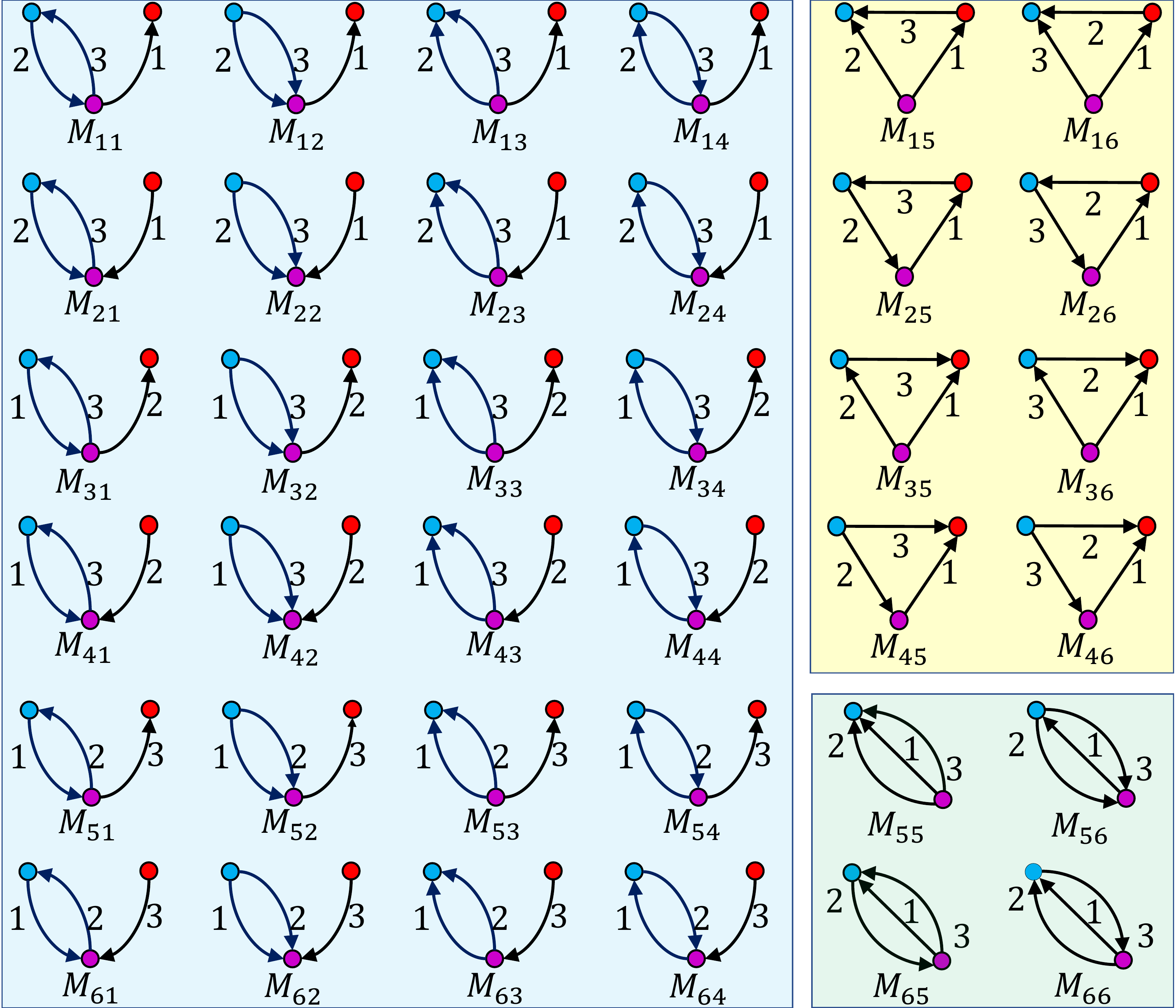}
\caption{All 2- and 3-node, 3-edge, $\delta$-temporal motifs.} 
\label{fig.TemporalMotifs}
\vspace{-3mm}
\end{center}
\end{figure}

In this paper, we focus on 2- and 3-node, 3-edge, $\delta$-temporal motifs, which are considered as the most common types of motifs~\cite{benson2016higher}. 
As shown in Fig.~\ref{fig.TemporalMotifs}, there are 32 kinds of 3-node, 3-edge, $\delta$-temporal motifs and 4 kinds of 2-node, 3-edge, $\delta$-temporal motifs. These 36 kinds of $\delta$-temporal motifs constitute the basic motifs in temporal graphs~\cite{paranjape2017motifs}.
In the toy example shown in Fig.~\ref{fig.toyexample}, the edge sequence $S=\langle (v_a,v_c,4s), (v_a,v_c,8s), (v_d,v_a,9s)\rangle$ is a motif instance of temporal motif $M_{63}$, $S=\langle (v_e,v_c,6s),  (v_d,v_c,10s), (v_d, v_e, \\ 14s)\rangle$ is a motif instance of temporal motif $M_{46}$, and $S=\langle (v_d,v_e,14s),(v_e,v_d,18s), (v_d,v_e,21s)\rangle$ is a motif instance of 2-node pair temporal motif $M_{65}$. 

And it also allows users to specify the types of motifs to their interest in Def. 2


\begin{Problem}[$\delta$-temporal Motif Counting]
Give a temporal graph $\mathcal{G}$ and a time interval $\delta$, temporal motif counting is to exactly count the number of motif instances for all 2-node and 3-node, 3-edge, $\delta$-temporal motifs in $\mathcal{G}$. 
\end{Problem}

Key notations used in the paper are summarized in Table~\ref{table_notations}. 

\begin{table}[htbp]
\centering
\vspace{-2mm}
\caption{Main notations and their definitions.}
\vspace{-2mm}
\label{table_notations}
\begin{tabular}{c|c}
     \toprule
     Notation & Definition \\
     \midrule
     $\mathcal{G}$			& input temporal network\\
     $\mathcal{V}, \mathcal{E}, \mathcal{T}$ & node/edge/timestamp set of $\mathcal{G}$\\
     $(v_i, v_j, t)$	& an edge from $v_i$ to $v_j$ with timestamp $t$\\
     $M_{ij}$	        & a $\delta$-temporal motif \\
     $u$                & center node \\
     $e=(t,v,dir)$      & an edge \wrt. current center node $u$ \\
     $e.t, e.v, e.dir$  & timestamp/node on the other side/direction of $e$\\
     $S_u$   & the edge sequence of center node $u$, sorted by time\\
     $Star[\cdot,\cdot,\cdot,\cdot]$   & the quadruple counter for star temporal motifs \\
     $Pair[\cdot,\cdot,\cdot]$   & the triple counter for pair temporal motifs \\
     $Tri[\cdot,\cdot,\cdot,\cdot]$   & the quadruple counter for triangle temporal motifs \\
     $\delta$			& the time constraint for $\delta$-temporal motif\\
     $thr_{d}$			& the degree threshold for hierarchical parallel framework\\
     \bottomrule
\end{tabular}
\vspace{-2mm}
\end{table}

\section{Methodology}



In this section, we propose two fast algorithms for exactly counting motif instances of all 2- and 3-node, 3 edge, $\delta$-temporal motifs in a given temporal graph. According to topological structures, we first divide all 2- and 3-node, 3-edge, $\delta$-temporal motifs (in Fig.~\ref{fig.TemporalMotifs}) into three categories: \textit{pair temporal motifs} (with green background), \textit{star temporal motifs} (with blue background) and \textit{triangle temporal motifs} (with yellow background). 
Pair temporal motifs only include 2 nodes with 3 temporal edges. 
Star and triangle temporal motifs contain 3 nodes with 3 temporal edges forming star structure and triangle structure respectively. 
Considering the structural similarity between pair temporal motifs and star temporal motifs, we use an unified fast exact algorithm to count both of them. 

\subsection{Proposed Method for Star and Pair Temporal Motifs}
In fact, there are 4 non-isomorphic pair temporal motifs and 24 non-isomorphic star temporal motifs (in Fig.~\ref{fig.TemporalMotifs}). 
We first focus on the 24 non-isomorphic star temporal motifs. In each kind of star temporal motif, we denote the node with the largest degree (connected with 3 edges) as the center node $u$. Each edge connected to center node $u$ in graph $\mathcal{G}$ can be defined by: (i) the timestamp $t$ of the edge, (ii) another node $v$ linked to the edge, and (iii) the direction $dir$ \wrt. center node $u$ (outward from or inward to $u$). That is, each edge connected to center node $u$ can be expressed as $e = (t, v, dir)$. 

\subsubsection{Three Types of Star Temporal Motifs}

\begin{figure}[h]
\begin{center}
\vspace{-2mm}
\includegraphics[width=1.0\linewidth]{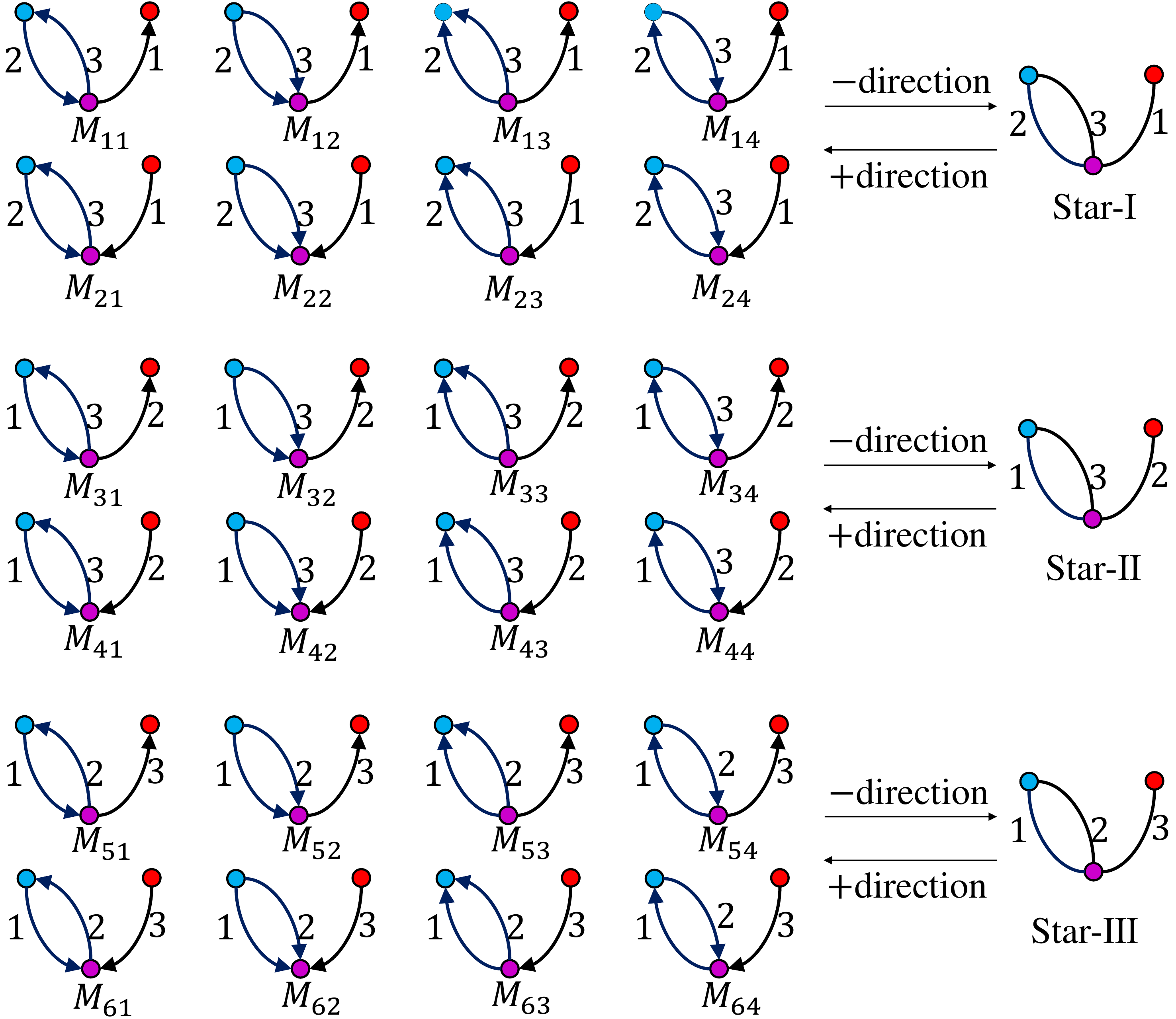}
\caption{Types of star temporal motifs.} 
\label{fig.StarMotifs}
\vspace{-3mm}
\end{center}
\end{figure}

Star temporal motifs are not centrosymmetric -- 2 edges both connect to two nodes while one isolated edge connects to another node. If we ignore the directions of edges, but only consider the time order of edges, we can divide these 24 kinds of star temporal motifs into three types based on the time order of the isolated edge (see Fig.~\ref{fig.StarMotifs}): 
\begin{itemize}
\item \textbf{Star-\uppercase\expandafter{\romannumeral1}}: The first edge in time order is isolated and connects to one node. The second and third edges both connect to another node.
\item \textbf{Star-\uppercase\expandafter{\romannumeral2}}: The second edge in time order is isolated and connects to one node. The first and third edges both connect to another node.
\item \textbf{Star-\uppercase\expandafter{\romannumeral3}}: The third edge in time order is isolated and connects to one node. The second and third edges both connect to another node.
\end{itemize}

\subsubsection{Quadruple Counter for Star Temporal Motifs}

We now introduce a quadruple counter $Star[Type, dir_1, dir_2, dir_3]$ for counting the number of motif instances of all star temporal motifs, which considers both motif type and directions of three edges. 
Specifically, the first dimension (\ie, $Type$) indicates the type of star temporal motif (\ie, Star-I, Star-II, or Star-III). The next three dimensions, \ie, $dir_1$, $dir_2$ and $dir_3$, represent the direction of first edge, second edge, and third edge in each motif type, respectively. 
Notice that each edge has two direction options: outward from or inward to its center node $u$, which is denoted by $o$ for outward or $in$ for inward respectively. 
Therefore, this quadruple counter can record $3 \times 2 \times 2 \times 2 = 24$ kinds of motifs, which correspond to 24 non-isomorphic star temporal motifs in Fig.~\ref{fig.TemporalMotifs}. 
For example, $Star[{\rm I},in,o,in]$ records the number of motif instances of $M_{24}$, because edge pattern of $M_{24}$ follows that of Star-${\rm I}$, and the first and third edges all link inward to the center node, while the second edge moves outward from center node.

\subsubsection{Fast Counting Algorithm for Star and Pair Temporal Motifs}

For counting all motif instances of all star and pair temporal motifs, we treat each node in the given temporal graph as the center node $u$ recursively.  
Specifically, for each node $u$, we list its all linked edges in time order as $S_u=\langle(t_1,v_1,dir_1),(t_2,v_2,dir_2),\dots,(t_s,v_s,dir_s)\rangle$, where $t_1\leq t_2\leq \dots \leq t_s$. 
We aim to detect all motif instances from the edge sequence $S_u$ \wrt. center node $u$. Intuitively, every edge in $S-\langle (t_{s-1},v_{s-1},dir_{s-1}), (t_s,v_s,dir_s)\rangle$ can be regraded as the first edge of a star temporal motif. 
Suppose that $(t_i,v_i,dir_i)$ is currently selected as the first edge for a star temporal motif, then $(t_j,v_j,dir_j)$ is selected as the third edge one by one from $t_{i}$ until $t_j - t_i \geq \delta$. 
In each process, by scanning all edges between the first edge and the third edge, the type of star temporal motif can be determined, and the corresponding number of motif instances can also be recorded. 

Next, we elaborate the details of our proposed \models algorithm for counting all star and pair temporal motifs. 
Given an edge sequence $S_u$ \wrt. the center node $u$, we now choose $e_1=(t_i, v, dir_i) \in S_u, e_3=(t_j, w, dir_j) \in S_u$ $(t_j-t_i \le \delta)$ as the first edge and third edge respectively. Let $e_2=(t_k, x, dir_k)\ (t_i\leq t_k \leq t_j)$ be the second edge candidate. 

\begin{figure}[h]
\begin{center}
\vspace{-3mm}
\includegraphics[width=1\linewidth]{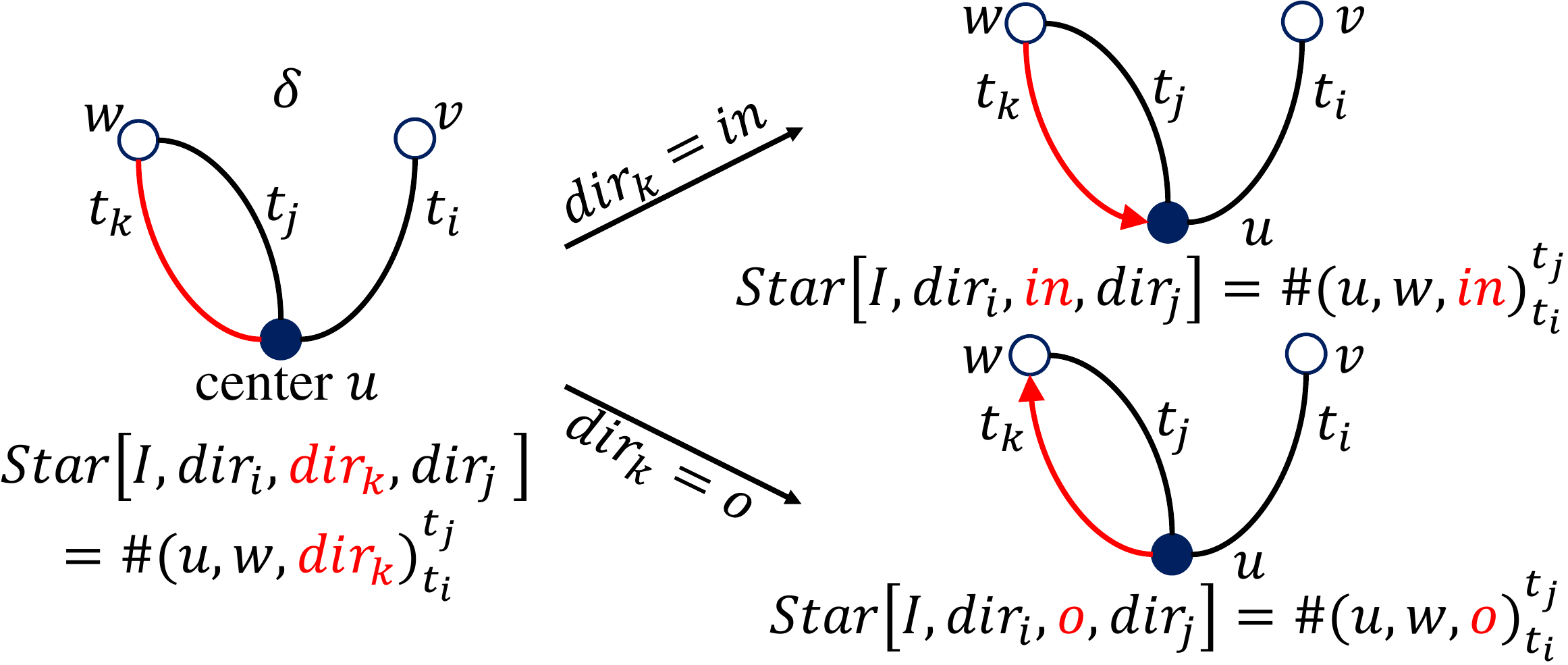}
\vspace{-3mm}
\caption{Counting Star-\uppercase\expandafter{\romannumeral1} motifs by the second edge.} 
\label{fig.StarMotifsCase1}
\vspace{-1mm}
\end{center}
\end{figure}

\begin{figure}[h]
\begin{center}
\vspace{-4mm}
\includegraphics[width=1\linewidth]{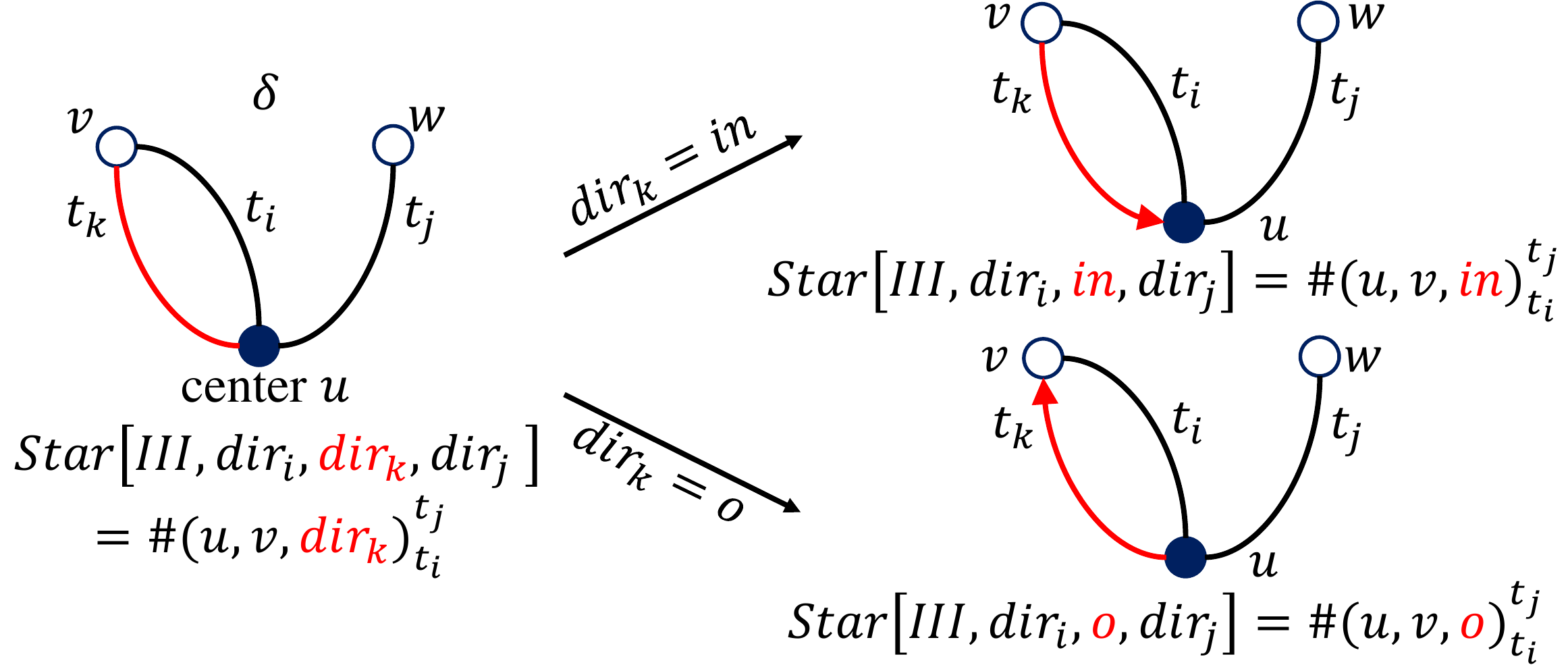}
\caption{Counting Star-\uppercase\expandafter{\romannumeral3} motifs by the second edge.} 
\label{fig.StarMotifsCase2}
\vspace{-4mm}
\end{center}
\end{figure}

We first discuss the different cases for temporal motif counting according to the choose of the second edge:  

(a) If $v \neq w$ (\ie, $e_1.v \neq e_3.v$), only the edges connected to $v$ or $w$ can be selected as the second edge candidates to form a star temporal motif. 
According to types of star temporal motifs shown in Fig.~\ref{fig.StarMotifs}, if $x=w$, then the formed motif belongs to \textbf{Star-I}, and if $x=v$, it belongs to \textbf{Star-III}. 
Specifically, the specific kind of star temporal motif depends on the exact directions of the three edges. 
The number of motif instances for each specific motif kind is equal to the number of second edge candidates with same $x$ and $dir_k$. 
As shown in Fig.~\ref{fig.StarMotifsCase1} and Fig.~\ref{fig.StarMotifsCase2}, the number of motif instances $Star[{\rm I/III},dir_i,dir_k,dir_j]$ is equal to the number of the second edge candidates $\#(u,w/v,dir_k)_{t_i}^{t_j}$,  
where $\#(u,w/v,dir_k)_{t_i}^{t_j}$ denotes the number of edges between center node $u$ and $w/v$ with direction $dir_k$ \wrt. $u$ and timestamp $t_k\in [t_i,t_j]$.

\begin{figure}[h]
\begin{center}
\vspace{-3mm}
\includegraphics[width=1\linewidth]{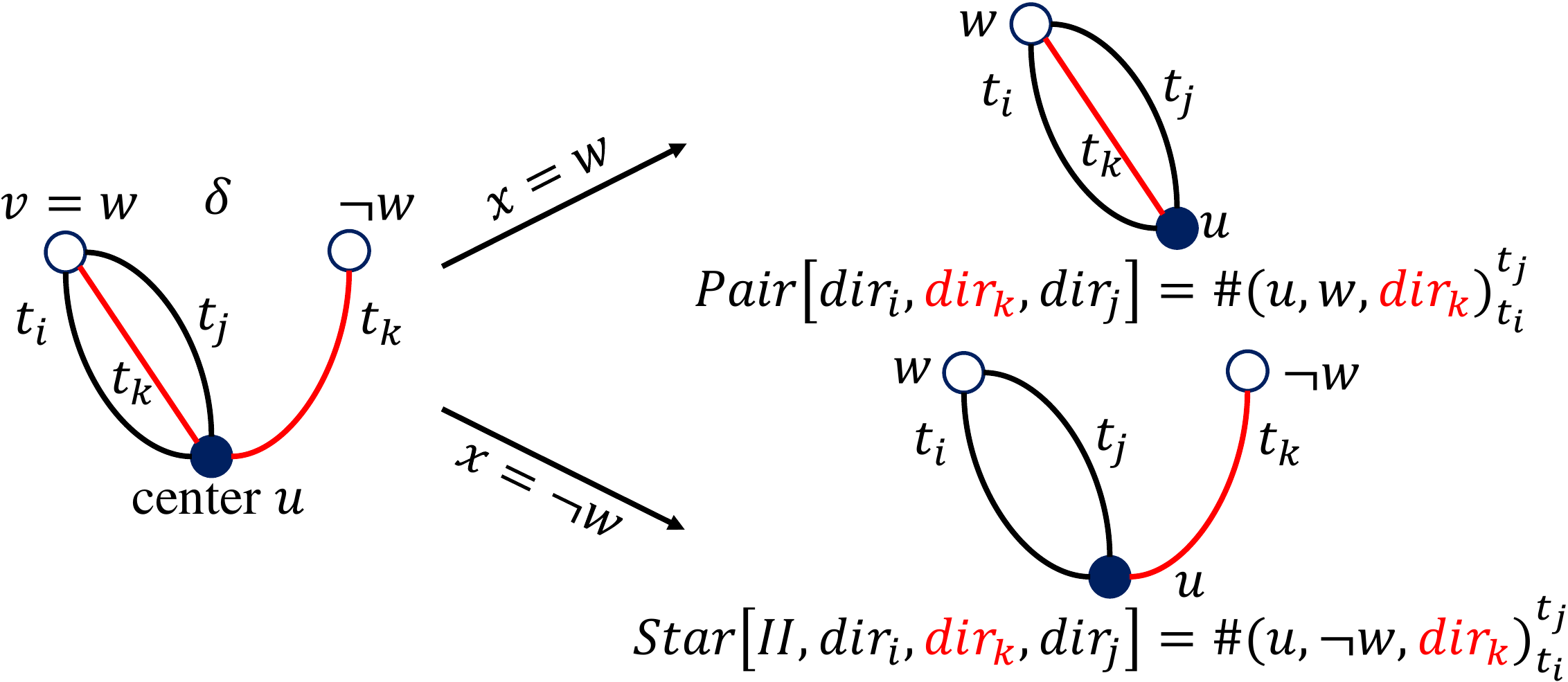}
\caption{Counting Star-\uppercase\expandafter{\romannumeral2} and Pair Temporal Motifs.} 
\label{fig.StarMotifsCase3}
\vspace{-6mm}
\end{center}
\end{figure}

(b) If $v = w$, all edges between the first edge and the third edge can be chose as the second edge to form a temporal motif. 
It is worth noting that when $x = w = v$, the currently formed motif is a \textbf{pair temporal motif}. 
In other cases, the formed motifs belong to \textbf{Star-II}. 
Similarly, the specific kind of pair/star temporal motif depends on the directions of the three edges, and the number of motif instances is equal to the number of second edge candidates. 
Following the counter $Star[Type, dir_1, dir_2, dir_3]$, we use a triple counter $Pair[dir_1, dir_2, dir_3]$ to record the number of motif instances for pair temporal motifs. 
As shown in Fig.~\ref{fig.StarMotifsCase3}, the number of pair motif instances $Pair[dir_i,dir_k,dir_j]$ is equal to the number of the second edge candidates $\#(u,w,dir_k)_{t_i}^{t_j}$,  
and the number of star motif instances $Star[{\rm{II}}, dir_i,dir_k,dir_j]$ is equal to the number of the second edge candidates $\#(u,\neg w,dir_k)_{t_i}^{t_j}$, 
where $\#(u,\neg w,dir_k)_{t_i}^{t_j}$ denotes the number of edges between node $u$ and other nodes except $w$ with direction $dir_k$ \wrt. $u$ and timestamp $t_k\in [t_i,t_j]$. 

Notice that there are only 4 kinds of non-isomorphic pair temporal motifs, while here we count the number of motif instances for 8 kinds of pair temporal motifs in counter $Pair[\cdot,\cdot,\cdot]$. This is because each pair of them are isomorphic, \ie, $Pair[in,in,in] \cong Pair[o,o,o] \cong M_{55}$, $Pair[in,o,o] \cong Pair[o,in,in] \cong M_{56}$, $Pair[in,o,in] \cong Pair[o,in,o] \cong M_{65}$, and $Pair[in,o,in] \cong Pair[o,in,o] \cong M_{66}$.

Taking the temporal graph shown in Fig.~\ref{fig.toyexample} as an example, suppose that $\delta =$ 10 seconds and node $v_a$ is selected as the center node $u$, the edge sequence of node $v_a$ is $S_a = \langle(4s,v_c,o),  (8s,v_c,o), (9s,v_d,in), (11s,v_b,o), (15s,v_c,o)\rangle$. 

First, let $e_1=S_a[1]=(4s,v_c,o)$ and $e_3=S_a[3]=(9s,v_d,in)$. So the second edge candidate can only choose $S_a[2]=(8s,v_c,o)$. Since $e_1.v\neq e_3.v$ (\ie, $v_c\neq v_d$) and $e_2.v=v_c=e_1.v$, the currently formed motif is a Star-III motif, hence $Star[{\rm III},o,o,in]$ += $1$. 
Next, $e_1$ remains unchanged, and let $e_3=S_a[4]=(11s,v_b,o)$. Since $e_1.v\neq e_3.v$, there is only one candidate for the second edge, \ie, $(8s,v_c,o)$. This is because only the edges connected to $e_1.v$ (\ie, $v_c$) or $e_3.v$ (\ie, $v_b$) can be selected as the second edge candidates to form 3-node star motif. According to the second edge ($(8s,v_c,o)$), the formed motif is also a Star-III motif, and thus $Star[{\rm III},o,o,o]$ += $1$. 
Then, $e_1$ remains unchanged, and let $e_3=S_a[5]=(15s,v_c,o)$. Because $15s-4s=11s>\delta$, $(15s,v_c,o)$ and the edges after it can no longer be chosen as the third edge \wrt. the current $e_1$. 

Therefore, we need to move the first edge $e_1$ to $S_a[2]=(8s,v_c,o)$, and select edge $S_a[4]=(11s,v_b,o)$ as the initial third edge $e_3$. The traversal process of the second edge is the same as above. Since edge $S_a[3]=(9s,v_d,in)$ does not meet the requirements of the node (\ie, $v_c$ or $v_b$), it can not constitute a motif with the current $e_1$ and $e_3$. 
Next, keep $e_1$ unchanged, and move $e_3$ to $S_a[5]=(15s,v_c,o)$. Since $e_1.v=e_3.v=v_c$, all edges between $e_1$ and $e_3$ can be selected as the second edge. Thus, the formed motifs by $(9s,v_d,in)$ and $(11s,v_b,o)$ are both Star-II motifs, and $Star[{\rm II},o,in,o]$ += $1$ and $Star[{\rm II},o,o,o]$ += $1$. 
So far, the star and pair temporal motif counting \wrt. the node $v_a$ is over. 

In this way, we treat each node as the center node in turn, and then we can count all motif instances for all star temporal motifs as well as pair temporal motifs.

\subsubsection{Complexity Analysis}
\label{sec.complexity}

Algorithm~\ref{alg.FastStar} shows the pseudo-code of the proposed \models for counting all kinds of star and pair temporal motifs. 


As depicted in lines 7-28 in Algorithm~\ref{alg.FastStar}, instead of traversing all edges between the first edge and the third edge for the choice of the second edge, we leverage the traversal of the third edge to realize the scan of the second edge, which significantly reduces the computation cost. 
More specifically, in the process of determining the third edge we use two HashMaps (\ie, $m_{in}$ and $m_{out}$) to simultaneously record the direction and the number of all the second edge candidates, and thus we no longer need to traverse all the edges again between the first and the third edges. 
Since $m_{in}$ and $m_{out}$ stores the number of edges connecting different nodes between the first and third edges ($m_{in}$ and $m_{out}$ respectively records two different directions relative to center node $u$), the number of the second edge candidates (\ie, $\#(u,v,in/o)_{t_i}^{t_j}$, $\#(u,w,in/o)_{t_i}^{t_j}$) can be queried in $m_{in}$ and $m_{out}$ only based on the first edge and the third edge. 
Notice that $m_{in}$ and $m_{out}$) always are initialized to be empty at the beginning of each iteration.

We now analyze the time complexity of our proposed \models for counting star and pair temporal motifs. We suppose that the degree of the $i$-th node in the temporal graph is $d_i$, and the average number of edges connected to the $i$-th node within time interval $\delta$ is $d_i^{\delta}$. 
For each node, the worst time complexity of the counting process is $O(d^{2})$. Therefore, the worst time complexity for whole graph is $O(\sum_{i=1}^{|\mathcal{V}|}(d_i^{2}))$. 

If we take time constraint $\delta$ into consideration, then the traversal space of the third edge is reduced to $d_i^{\delta}$. Therefore, the time complexity of counting process for $i$-th node becomes $O(d_i d_i^{\delta})$, and the worst time complexity for whole graph is $O(\sum_{i=1}^{|\mathcal{V}|}(d_i d_i^{\delta}))$. 
Since $d_i^{\delta} << d_i$, assuming that $d_i^{\delta}$ of all nodes is approximately equal, denoted by $d^{\delta}$, thus $O(\sum_{i=1}^{|\mathcal{V}|}(d_i d_i^{\delta})) \approx O(d^{\delta}\sum_{i=1}^{|\mathcal{V}|}(d_i)) \approx O(2 d^{\delta} |\mathcal{E}|)$, where $|\mathcal{E}|$ is the number of temporal edges in graph $\mathcal{G}$. 
Namely, our \models algorithm achieves the time complexity linear in the number of temporal edges of the input graph.

\begin{algorithm}[t]
\renewcommand{\algorithmicrequire}{\textbf{Input:}}
\renewcommand{\algorithmicensure}{\textbf{Output:}}
\caption{\model algorithm for Star/Pair Temporal Motifs }
\label{alg.FastStar}
\begin{algorithmic}[1]
\REQUIRE Temporal graph $\mathcal{G}=\{\mathcal{V},\mathcal{E},\mathcal{T}\}$, and time constraint $\delta$.
\ENSURE Star counter $Star[\cdot,\cdot,\cdot,\cdot]$, pair counter $Pair[\cdot,\cdot,\cdot]$.
\FOR{each node $u\in \mathcal{V}$}
    \STATE get $S_u=\langle(t_1,v_1,dir_1),(t_2,v_2,dir_2),\dots,(t_s,v_s,dir_s)\rangle$;
    \FOR{$i=1$ to $s-2$}
        \STATE $e_1 \leftarrow S_u[i]$;
        \STATE $\#e_{in} \leftarrow 0$; $\#e_{out} \leftarrow 0$;
        \STATE $m_{in}, m_{out}$ $\leftarrow$ $HashMap.Init()$; 
        \FOR{$j=i+1$ to $s$}
            \STATE $e_3 \leftarrow S_u[j]$;
            \IF{ $e_3.t - e_1.t > \delta$}
                \STATE break;
            \ENDIF
            \IF{ $e_3.v == e_1.v $}
                \STATE $Pair[dir_i,in,dir_j]$ += $m_{in}[e_1.v]$;
                \STATE $Pair[dir_i,o,dir_j]$ += $m_{out}[e_1.v]$;
                \STATE $Star[{\rm II},dir_i,in,dir_j]$ += $\#e_{in} - m_{in}[e_1.v]$;
                \STATE $Star[{\rm II},dir_i,o,dir_j]$ += $\#e_{out} - m_{out}[e_1.v]$;
            \ELSE
                \STATE $Star[{\rm I},dir_i,in,dir_j]$ += $m_{in}[e_3.v])$;
                \STATE $Star[{\rm I},dir_i,o,dir_j]$ += $m_{out}[e_3.v])$;
                \STATE $Star[{\rm III},dir_i,in,dir_j]$ += $m_{in}[e_1.v])$;
                \STATE $Star[{\rm III},dir_i,o,dir_j]$ += $m_{out}[e_1.v])$;
            \ENDIF
            \IF {$e_3.dir == i$}
                \STATE $m_{in}[e_3.v]$ += $1$; $\#e_{in}$ += $1$;
            \ELSE
                \STATE $m_{out}[e_3.v]$ += $1$; $\#e_{out}$ += $1$;
            \ENDIF
        \ENDFOR
    \ENDFOR
\ENDFOR
\RETURN $Star[\cdot,\cdot,\cdot,\cdot], Pair[\cdot,\cdot,\cdot]$;
\end{algorithmic}
\vspace{-1mm}
\end{algorithm}
\vspace{-2mm}

\begin{figure*}[h]
    \centering
    \vspace{-6mm}
    \hspace{-6mm}
    \subfigure[Triangle-\uppercase\expandafter{\romannumeral1}]{
    \includegraphics[width=0.33\linewidth]{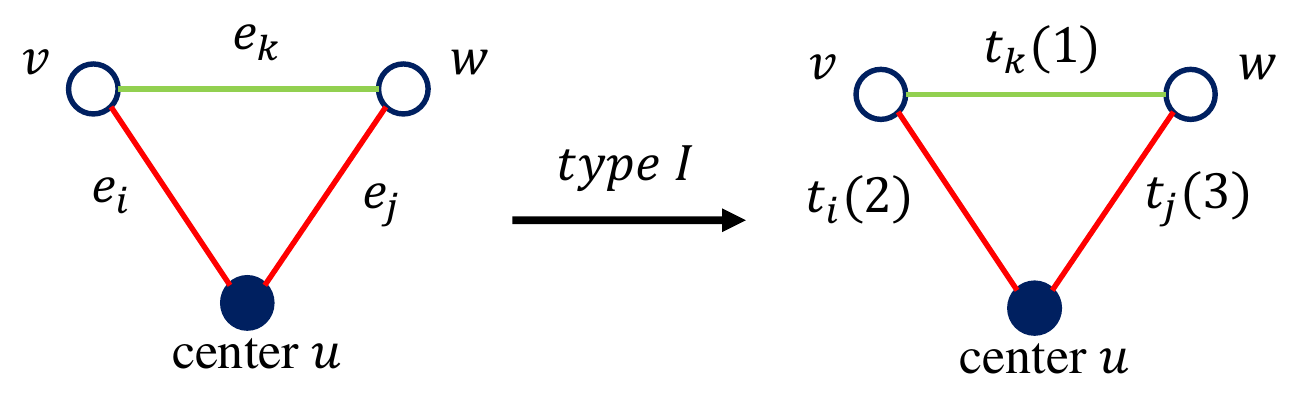}
    }
    \hspace{-3mm}
    \subfigure[Triangle-\uppercase\expandafter{\romannumeral2}]{
    \includegraphics[width=0.33\linewidth]{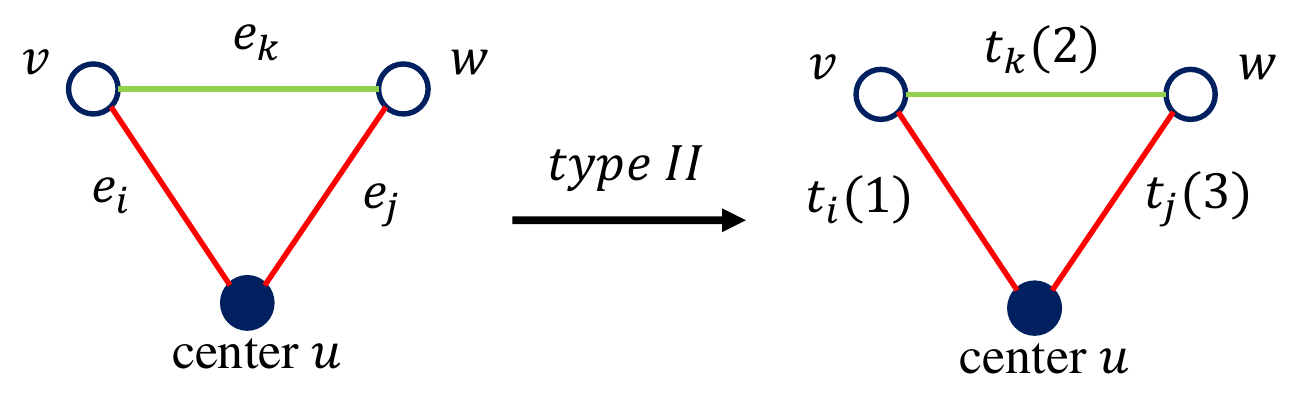}
    }
    \hspace{-3mm}
    \subfigure[Triangle-\uppercase\expandafter{\romannumeral3}]{
    \includegraphics[width=0.33\linewidth]{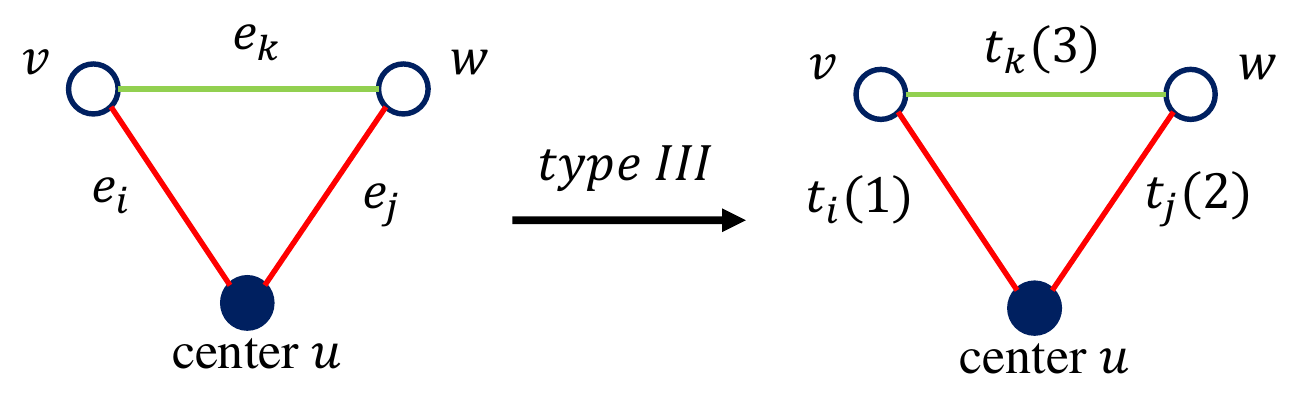}
    }
    \vspace{-3mm}
    \caption{Types of triangle temporal motifs.}
    \label{fig.TriangleType}
    \vspace{-3mm}
\end{figure*}

\subsection{Proposed Method for Triangle Temporal Motifs} 
\label{sec.triangle}

In this section, we present the details of our proposed exact counting algorithm \modelt for counting the number of motif instances for all triangle temporal motifs.

\subsubsection{Three Types of Triangle Temporal Motifs}


To identify potential triangle motifs, we first choose a node in graph $\mathcal{G}$ as the center node $u$, and then determine two different temporal edges connected to center node $u$ such that the other two connected nodes are different. We denote these two temporal edges as $e_i=(t_i,v,dir_i)$ and $e_j=(t_j,w,dir_j)$, where $t_j- t_i \leq \delta$, and $dir_i$ and $dir_j$ are the directions of $e_i$ and $e_j$ \wrt. center node $u$, respectively.  
Each edge between $v$ and $w$ and the two edges above may form triangle temporal motifs, depending on whether the time constraint $\delta$ is satisfied. 
Therefore, we next consider each edge $e_k$ between nodes $v$ and $w$ that satisfy the time constrain $\delta$. Let $t_k$ be the timestamp of edge $e_k$ and $dir_k$ be the direction of edge $e_k$ \wrt. node $v$ (\ie, $o$ indicates from $v$ to $w$ and $in$ denotes from $w$ to $v$). Unlike FAST-Star, the order of three edges is determined by their timestamps. Then according to the order of three edges, we determine the types of triangle temporal motif and use corresponding counter to count the triangle temporal motif.


\vspace{-0mm}
According to the time order of three edges (\ie, $t_k$, $t_i$ and $t_j$), we divide all kinds of triangle temporal motifs into three types (see Fig.~\ref{fig.TriangleType}): 

\begin{itemize}
    \item \textbf{Triangle-I}: If $t_k < t_i$ and $t_j- t_k \leq \delta$, then the formed triangle temporal motif belongs to Triangle-I type motif. 
    \item \textbf{Triangle-II}: If $t_i\leq t_k\leq t_j$, then the formed triangle temporal motif belongs to Triangle-II type motif. 
    \item \textbf{Triangle-III}: If $t_j < t_k$ and $t_k- t_i \leq \delta$, then the formed triangle temporal motif belongs to Triangle-III type motif. 
\end{itemize}

\vspace{-0mm}
\subsubsection{Fast Counting Algorithm for Triangle Temporal Motifs}

Considering the directions of the three edges for each type of triangle temporal motifs, we also introduce a quadruple counter $Tri[Type, dir_{i}, dir_{j}, dir_{k}]$ for counting the number of motif instances for all types of triangle temporal motifs. 
Notice that there are $3 \times 2 \times 2 \times 2 = 24$ counters in $Tri[Type, dir_{i}, dir_{j}, dir_{k}]$, while we have only 8 non-isomorphic triangle temporal motifs (see Fig.~\ref{fig.TemporalMotifs}). 
Namely, there exit some triangle temporal motifs recorded in $Tri[Type, dir_{i}, dir_{j}, dir_{k}]$ are isomorphic. 

\begin{figure}[h]
\begin{center}
\vspace{-3mm}
\includegraphics[width=1.0\linewidth]{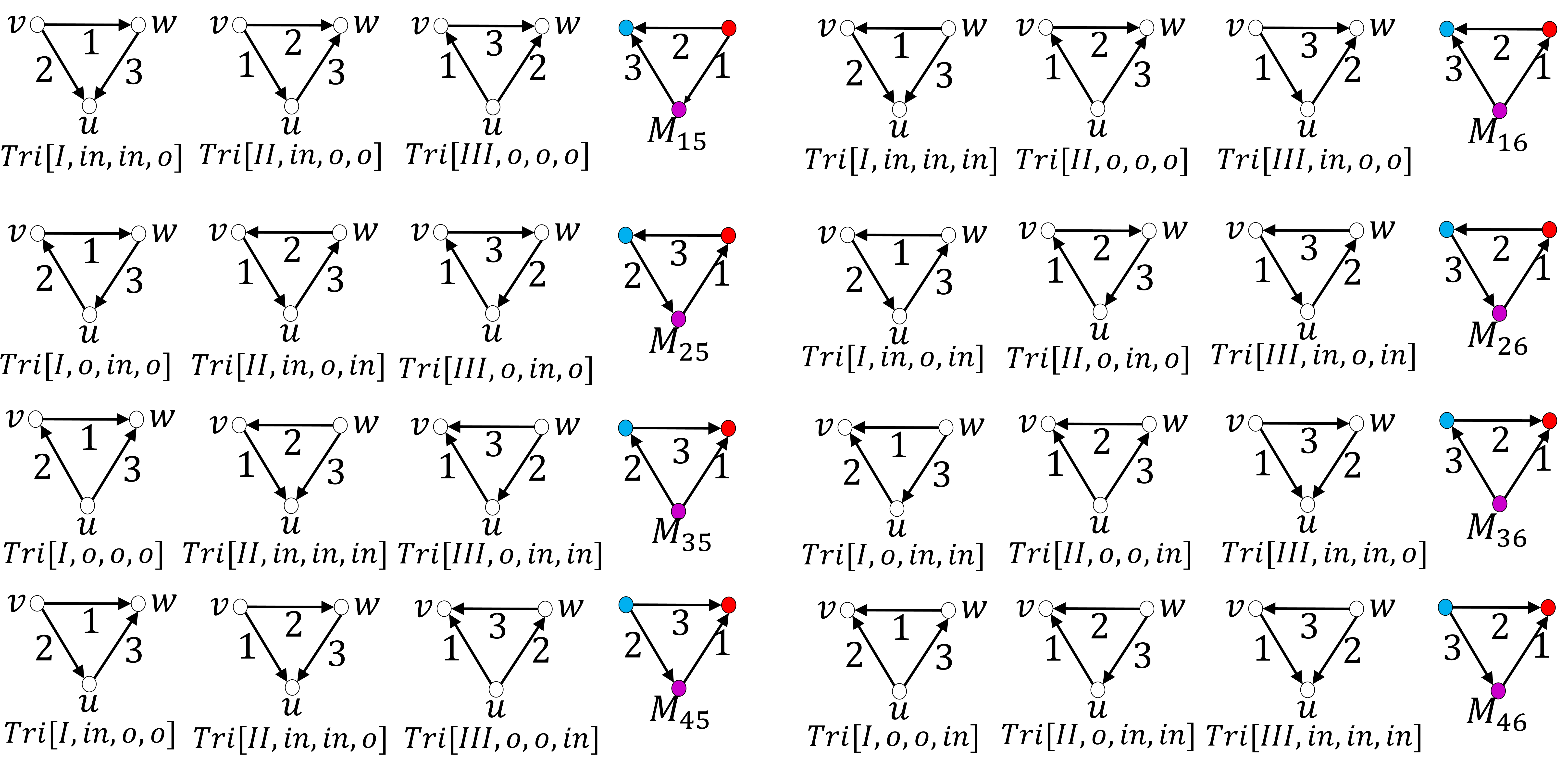}
\caption{Illustration of isomorphic triangle temporal motifs.}
\label{fig.TriangleIsomorphic}
\vspace{-7mm}
\end{center}
\end{figure}

As shown in Fig.~\ref{fig.TriangleIsomorphic}, the triangle temporal motifs corresponding to every three counters in $Tri[Type, dir_{i}, dir_{j}, dir_{k}]$ are isomorphic. Therefore, we only need to merge these isomorphic counters at the end for outputting the final result. 


Next, we use a specific example to illustrate how our algorithm counts triangle temporal motifs with time constraint $\delta$=$10$ seconds. 
Suppose that the current center node is node $e$ in the temporal graph shown in Fig.~\ref{fig.toyexample}. 
We already have the edge sequence $S_e=\langle(1s,v_d,o),  (6s,v_c,o), (14s,v_d,in), (18s,v_d,\\o), (21s,v_d,in)\rangle$ when counting the star/pair temporal motifs. 

We first select $S_e[1]$ (\ie, $(1s,v_d,o)$) as $e_i$, thus edge $e_j$ should meet two constraints: (i) $t_j-t_i \leq \delta$, and (ii) $e_j.v \neq e_i.v$. Therefore, only $S_e[2]=(6s,v_c,o)$ is qualified candidate for $e_j$. 
If $S_e[2]$ is selected as $e_j$, we then get the edge set between $v_c$ and $v_d$: $E_{(v_c, v_d)}=\{(v_d, v_c, 10s),(v_c, v_d, 17s)\}$. Since $10s-t_i=9s<\delta$ and $17s-t_i=16s > \delta$, only $(v_d, v_c, 10s)$ can be selected as $e_k$ to form a Triangle-III motif, and thus $Tri[{\rm III}, o, o, o]$ += $1$. 

Next, $e_i$ moves to $S_e[2]$, thus only $S_e[3]=(14s,v_d,in)$ is qualified candidate for $e_j$. If $e_j=S_e[3]$, only edge $(v_d, v_c, \\ 10s) \in E_{(v_c, v_d)}$ can be chosen as $e_k$ to form a Triangle-II motif, and thus $Tri[{\rm II}, o, in, o]$ += $1$.  
Among the remaining edges, no edge can be selected for $e_i$, because there is no qualified $e_j$ to be selected for each potential $e_i$. 
Now we find out all triangle temporal motifs with node $v_e$ as center node.

\subsubsection{Complexity Analysis}
\label{sec.tri-complexity}

Algorithm~\ref{alg.FastTriangle} shows the pseudo-code of our proposed \modelt for counting all triangle temporal motifs. 

\begin{algorithm}[t]
\renewcommand{\algorithmicrequire}{\textbf{Input:}}
\renewcommand{\algorithmicensure}{\textbf{Output:}}
\caption{\model algorithm for Triangle Temporal Motifs }
\label{alg.FastTriangle}
\begin{algorithmic}[1]
\REQUIRE Temporal graph $\mathcal{G}=\{\mathcal{V},\mathcal{E},\mathcal{T}\}$, and time constraint $\delta$.
\ENSURE Triangle counter $Tri[\cdot,\cdot,\cdot,\cdot]$. 
\FOR{each node $u\in \mathcal{V}$}
    \STATE get $S_u=\langle(t_1,v_1,dir_1),(t_2,v_2,dir_2),\dots,(t_s,v_s,dir_s)\rangle$;
    \FOR{$i=1$ to $s-1$}
        \STATE $e_i \leftarrow S_u[i]$;
        \FOR{$j=i+1$ to $s$}
            \STATE $e_j \leftarrow S_u[j]$;
            \IF{$e_j.t > e_i.t + \delta$ }
                \STATE break;
            \ENDIF
            \IF{$e_j.v==e_i.v$ }
                \STATE continue;
            \ENDIF
            \FOR{each edge $e \in E_{(e_i.v,e_j.v)}$} 
                \IF{$t_j - \delta \leq e.t < t_i$}
                    \STATE $Tri[{\rm I},dir_i, dir_j, e.dir]++$;
                \ELSIF{$t_i \leq e.t \leq t_j$ }
                    \STATE $Tri[{\rm II},dir_i, dir_j, e.dir]++$;
                \ELSIF{$t_j < e.t \leq t_i + \delta$}
                    \STATE $Tri[{\rm III},dir_i, dir_j, e.dir]++$;
                \ELSIF{$e.t > t_i + \delta$}
                    \STATE break;
                \ENDIF
            \ENDFOR
        \ENDFOR
    \ENDFOR
    \STATE $\mathcal{V} \leftarrow \mathcal{V}-\{u\}$; $\mathcal{E} \leftarrow \mathcal{E}-\{e_u\}$; //Only for single threading 
\ENDFOR
\RETURN $Tri[\cdot,\cdot,\cdot,\cdot]$
\end{algorithmic}
\end{algorithm}


Notice that each triangle temporal motif instance is counted three times repeatedly, because each instance can form three different types of triangle temporal motifs \wrt. its three vertices, respectively. 
For example, edge sequence $\langle(v_a, v_c, 8s), (v_d, v_a, 9s), (v_c, v_d, 17s)\rangle$ in Fig.~\ref{fig.toyexample} forms an instance of triangle temporal motif $M_{25}$. When $v_a$ is selected as the center node, this instance can be identified as an instance recorded in $Tri[{\rm III},o,in,o]$. When $v_c$ is selected as the center node, this instance will be detected as an instance recorded in $Tri[{\rm II},in,o,in]$. When $v_d$ is selected as the center node, this instance is recognized as an instance recorded in $Tri[{\rm I},o,in,o]$. 
In fact, the counters $Tri[{\rm III},o,in,o]$, $Tri[{\rm II},in,o,in]$ and $Tri[{\rm I},o,in,o]$ are isomorphic. Therefore, as shown in Fig.~\ref{fig.TriangleIsomorphic}, each triangle temporal motif can be identified as three distinct triangle types based on its three different vertices respectively. That is, each triangle temporal motif is counted three times.

However, it is easy to handle this issue in a single thread. That is, when the number of motif instances \wrt. center node $u$ has been counted, center node $u$ is removed from $\mathcal{V}$, including all connected edges, to avoid redundant counting. 
In this work, we target on a counting framework that is natively parallel. To achieve this, we have to avoid any dependency among different processes/threads, while eliminating the repeated counting will inevitably introduce dependency. 
Therefore, in multi-threading environment, to avoid any dependency among different processes/threads, we do not do any processing, but repeatedly count and finally divide by three.

We then analyze the time complexity of \modelt algorithm. Similarly, we use $d_i$ to denote the degree of the $i$-th node and $d_i^{\delta}$ to represent the average number of edges within $\delta$ for the $i$-th node. 
For each center node, the worst time complexity of searching two edges for constructing potential triangle motifs is $O(d_i d_i^{\delta})$. In the worst case, every combination of two edges can form a potential triangle motif. Hence, the worst time complexity for one node is $O(d_i d_i^{\delta} \xi)$, where $\xi$ denotes the average number of edges between two nodes. 
For the whole graph, the worst time complexity is $O(\sum_{i=1}^{|\mathcal{V}|} d_i d_i^{\delta} \xi)$. 
Using some implementation tricks, $\xi$ can be reduced to the number of edges between two nodes within $\delta$ time interval, \ie, $\xi \leq d_i^{\delta}$.  
Therefore, the worst time complexity for the whole graph is less than $O(\sum_{i=1}^{|\mathcal{V}|} d_i (d_i^{\delta})^2)$. 

Because $d_i^{\delta} << d_i$, we assume that $d_i^{\delta}$ of all nodes is approximately equal, denoted by $d^{\delta}$, thus $O(\sum_{i=1}^{|\mathcal{V}|} d_i (d_i^{\delta})^2) \approx O((d^{\delta})^2\sum_{i=1}^{|\mathcal{V}|}(d_i)) \approx O(2 (d^{\delta})^2 |\mathcal{E}|)$. 
That is, our \modelt also achieves linear time complexity with the input graph.

\subsection{Hierarchical Parallel Framework}
\label{sec.parallel}

\begin{figure}[h]
    \centering
    \vspace{-2mm}
    \hspace{-5mm}
    \subfigure[Degree distribution of nodes]{
    \includegraphics[width=0.49\linewidth]{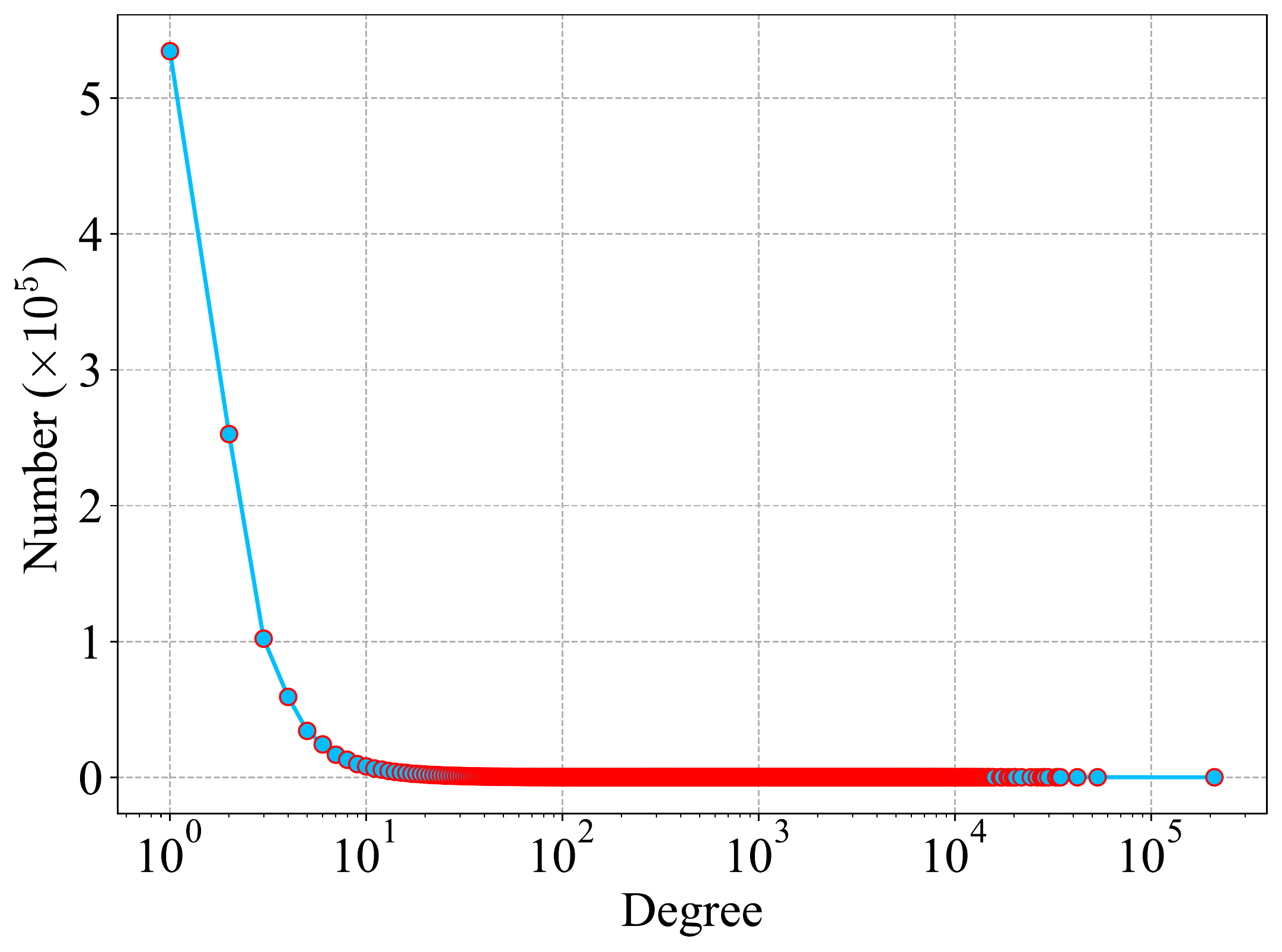}
    
    }
    \hspace{-3mm}
    \subfigure[Average run-time per node]{
    \includegraphics[width=0.49\linewidth]{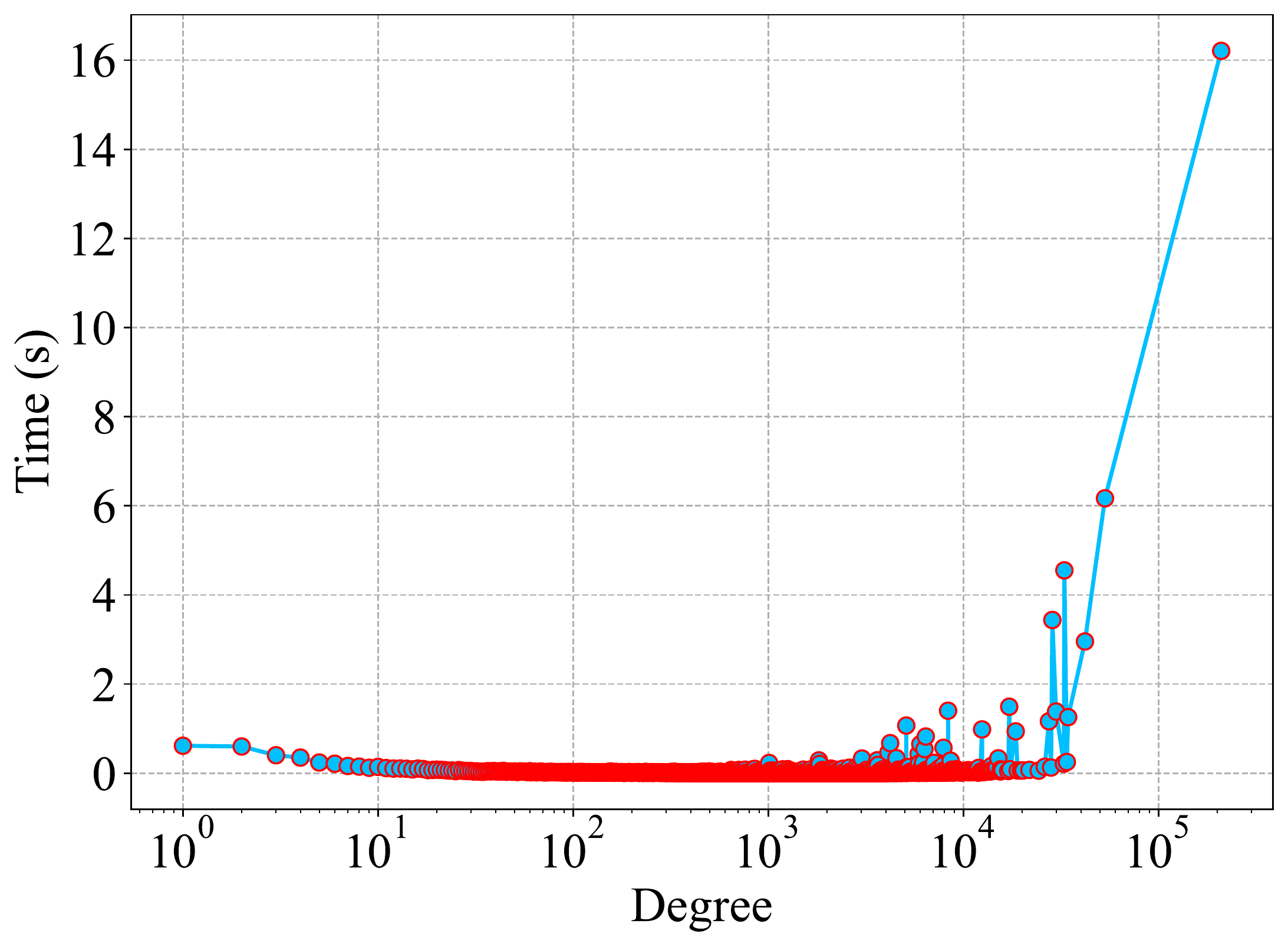}
    }
    \vspace{-2mm}
    \caption{Data statistics on WikiTalk.}
    \label{fig.degree}
    \vspace{-3mm}
\end{figure}

As described in Algorithm~\ref{alg.FastStar} and Algorithm~\ref{alg.FastTriangle}, our algorithms recursively tread each node in temporal graph as center node to detect all motif instances, which have no direct data dependency. Therefore, our proposed \model (general term for \models and \modelt) naturally has high parallelism, namely, our \model converts the temporal motif counting into an embarrassingly parallel problem. 
Nevertheless, we find that simply employing multi-threading do not achieve the desired effect (\eg, approximately linear speedup) on same graph datasets. 
This is because the degree distribution of the nodes in most temporal graphs is extremely unbalanced, which leads to the load unbalanced problem of multi-threading. Even if some balanced load strategies are considered at the node level (\eg, dynamic schedule), the expected results cannot be achieved. 
Fig.~\ref{fig.degree} shows the degree distribution of all nodes in \textit{WikiTalk} graph, and the average time consumption of counting all motif instances required per node with corresponding degree. 
We observe that although the degree distribution of the graph has a typical long-tailed distribution, the few nodes with higher degrees (\eg, top ten node) account for the dominant part of the time consumption of whole graph. 
Fortunately, from Algorithm~\ref{alg.FastStar} and Algorithm~\ref{alg.FastTriangle}, it is not difficult to see that temporal motif counting process inside the node in our \model also has high parallelism. 

To address the above issue, we propose a Hierarchical pArallel fRamEwork (\framework) for the proposed \model to accelerate the counting of temporal motifs powerfully. 
Specifically, our framework consists of two parallel strategies: \textit{inter-node parallel} and \textit{intra-node parallel} mechanisms. 
More specifically, since our proposed \model has inherent parallelism both within and among nodes, we implement these two parallel mechanisms for our \framework based on OpenMP. 

\textit{Inter-node parallel}: because the temporal motif counting of different nodes are independent of each other, we can assign one or more nodes to a thread, and assign all nodes to different threads to realize parallel motif counting for all nodes. In addition, we use the dynamic scheduling mode provided by OpenMP to effectively solve the problem of load imbalance. 

\textit{Intra-node parallel}: inside each node, the motif counting process can be divided into subtasks according to different starting edges. In fact, data reading and writing operations within the node are more vulnerable to multi-threading, which can result in unpredictable results. To ensure parallelism within the node, we use the reduction tool provided by OpenMP to copy the variables within node, so that each thread keeps the backup of these variables, and then reduce and output the final result after all tasks are completed. 

In particular, we set a degree threshold $thr_d$ for our hierarchical parallel framework. 
For the nodes with a degree greater than $thr_d$, we will enable intra-node parallel mode, otherwise, perform inter-node parallel strategy.


It is worth noting that HARE is able to directly use OpenMP tool to handle load imbalance in both inter-node and intra-node parallel modes, because we design HARE to be natively parallel. There is no any dependency among different threads.


\section{Experiment}


\subsection{Datasets}
We conduct extensive experiments on sixteen real-world temporal networks, which are available publicly in~\cite{snapnets}\cite{nr}. 

\begin{table}[t]
\begin{center}
\vspace{-2mm}
\caption{Basic statistics of twelve temporal networks.}
\vspace{-1mm}
\label{table_dataset}
\setlength{\tabcolsep}{2.8mm}{}
\begin{tabularx}{\linewidth}{c | c | c | c}
\toprule
Dataset & \#nodes & \#temporal edges & Time span (day) \\
\midrule
Email-Eu & 986 & 332,334 & 803\\
CollegeMsg & 1,899  & 20,296 & 193\\
Bitcoinotc & 5,881 & 35,592 & 1,903\\
Bitcoinalpha & 3,783 & 24,186 & 1,901 \\
Act-mooc & 7,143 & 411,749 & 29 \\
SMA-A & 44,090 & 544,817 & 338 \\
FBWALL & 45,813 & 855,542 & 1,591 \\
MathOverflow & 24,818 & 506,550 & 2,350 \\
AskUbuntu & 159,316 & 964,437 & 2,613 \\
SuperUser & 194,085 & 1,443,339 & 2,773 \\
Rec-MovieLens & 283,228 & 27,753,444 & 1,128 \\
WikiTalk & 1,140,149 & 7,833,140 & 2,320 \\
StackOverflow & 2,601,977 & 63,497,050 & 2,774 \\
IA-online-ads & 15,336,555 & 15,995,634 & 2,461 \\
Soc-bitcoin & 24,575,382 & 122,948,162 & 2,584 \\
RedditComments & 8,036,164 & 613,289,746 & 3,686 \\
\bottomrule
\end{tabularx}
\vspace{-5mm}
\end{center}
\end{table}

\begin{figure*}[h]
    \centering
    \vspace{-4mm}
    \subfigure[CollegeMsg]{
    \includegraphics[width=0.47\linewidth]{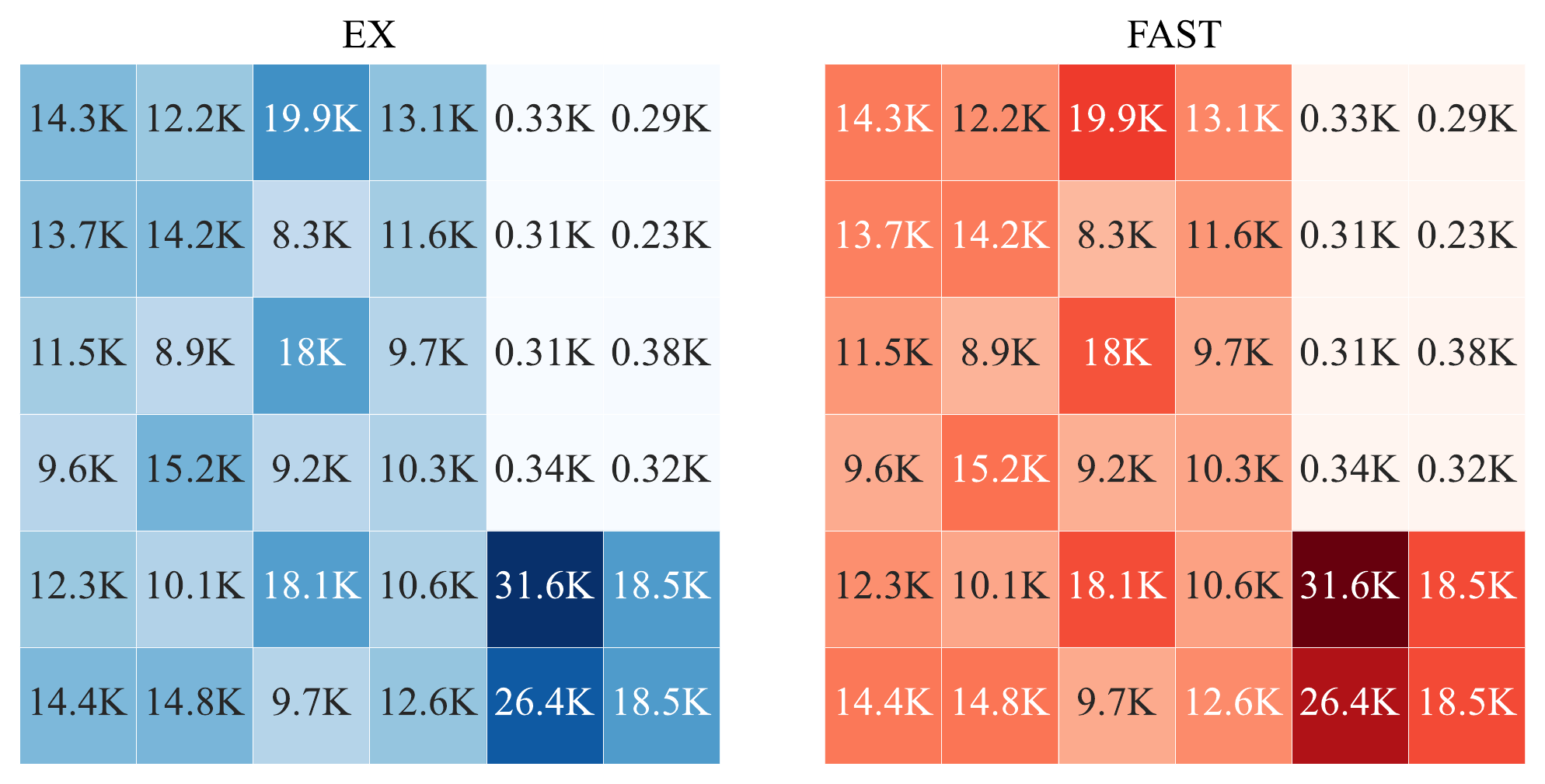}
    }
   \hspace{1mm}
    \subfigure[Superuser]{
    \includegraphics[width=0.47\linewidth]{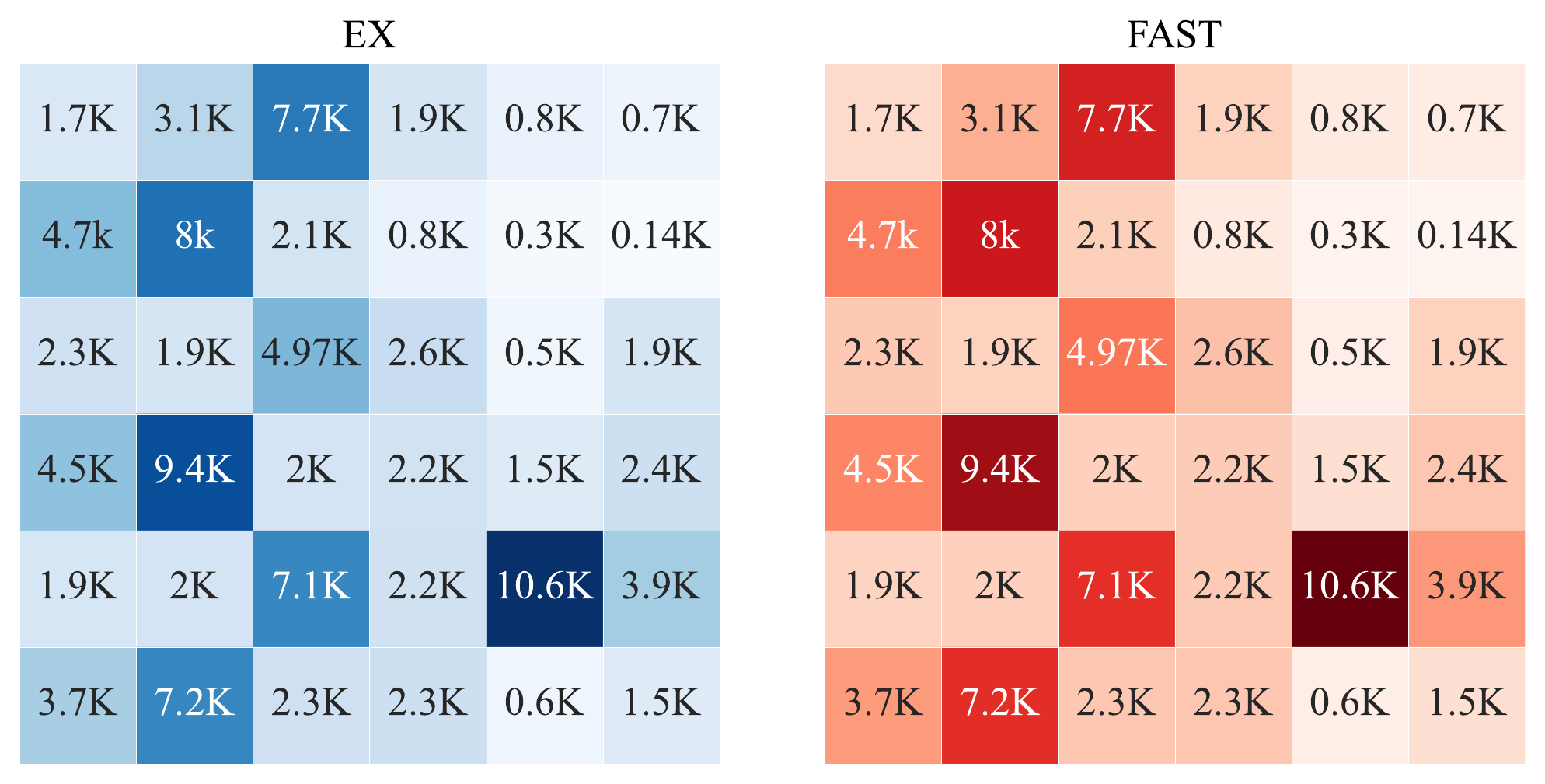}
    }
    \\
    \vspace{-2mm}
   \subfigure[WikiTalk]{
    \includegraphics[width=0.47\linewidth]{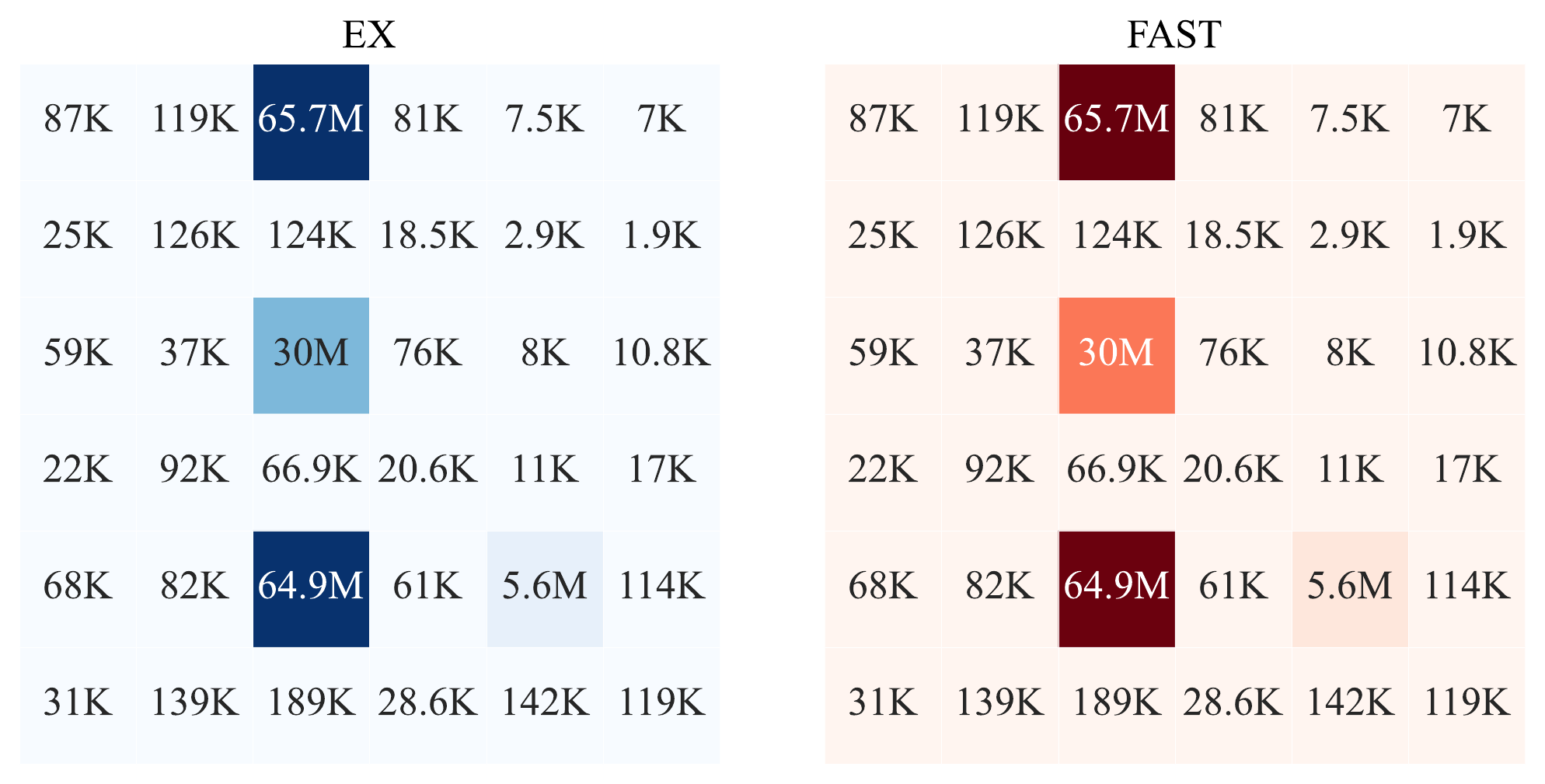}
    }
    \hspace{1mm}
    \subfigure[StackOverflow]{
    \includegraphics[width=0.47\linewidth]{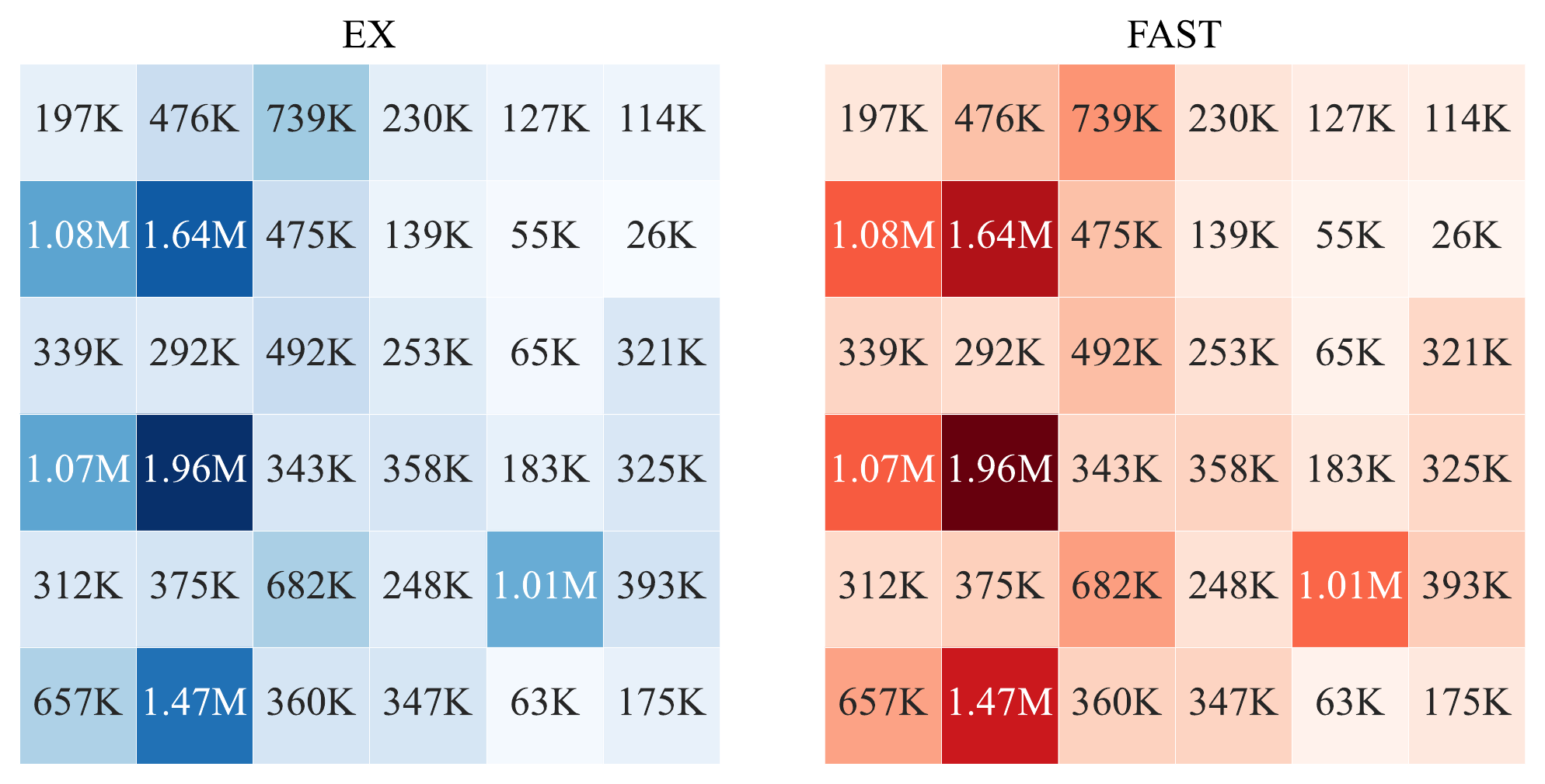}
    }
    \vspace{-2mm}
    \caption{Counts of motif instances of all 2-node and 3-node, 3-edge $\delta$-temporal motifs with $\delta=600s$. For each dataset, counts in the $i$-th row and $j$-th column is the number of motif instances of $M_{ij}$. The color for motif $M_{ij}$ indicates the fraction over all $M_{ij}$ on a linear scale -- darker blue/red means a higher count.}
    \label{fig.instances}
    \vspace{-3mm}
\end{figure*}

\begin{itemize}
    \item \textbf{Email-Eu} is a collection of internal email records from a European research institution. An edge $(u, v, t)$ signifies that person $u$ sent person $v$ an email at time $t$. 
    
    \item \textbf{CollegeMsg} is a network of private messages sent on an online social network at the University of California, Irvine. 
    
    \item \textbf{Bitcoinotc} and \textbf{Bitcoinalpha} are Bitcoin OTC and Bitcoin Alpha web of trust network respectively. 
    An edge $(u, v, t)$ in these datasets denotes that bitcoin was transferred from address $u$ to address $v$ at time $t$.
    
    \item \textbf{Act-mooc} is a collection of student actions on a popular MOOC platform. The actions are represented as a directed, temporal network. 
    
    \item \textbf{SMS-A} is a texting service provided on mobile phones. In this dataset, an edge $(u, v, t)$ means that person $u$ sent an SMS message to person $v$ at time $t$. 
    
    \item \textbf{FBWALL} is derived from the social network Facebook located in the New Orleans region, where the edges are wall posts between users. 
    
    \item \textbf{MathOverflow}, \textbf{Askubuntu}, \textbf{Superuser} and \textbf{StackOverflow} are derived from user interactions on Stack Exchange question and answer forums. A temporal edge represents a user replying to a question, replying to a comment, or commenting on a question.  
    
    \item \textbf{Rec-MovieLens} is a rating dataset from the MovieLens website, where an edge $(u, v, t)$ signifies that user $u$ rated the movie $v$ at time $t$.  
    
    \item \textbf{WikiTalk} is a network of Wikipedia users making edits on each others' ``talk pages''. An edge $(u, v, t)$ indicates that user $u$ edited user $v'$s talk page at time $t$. 
    
    \item \textbf{IA-online-ads} contains information about the product related advertisements a user has clicked. An edge $(u, v, t)$ indicates that user $u$ clicked on advertisement $v$ at $t$.
    
    
    \item \textbf{Soc-bitcoin} is a large-scale bitcoin transaction network where each edge $(u, v, t)$ denotes that bitcoin was transferred from address $u$ to address $v$ at time $t$.

    \item \textbf{RedditComments} is constructed from a large collection of comments made by users on a popular social media platform \url{https://www.reddit.com}. An edge $(u, v, t)$ indicates that a comment from user $u$ to user $v$ at time $t$. 
\end{itemize}

Specifically, we collect seven large-scale datasets with more than one millions of temporal edges to better analyze the performance of our framework, especially, Soc-bitcoin and RedditComments have more than 100M temporal edges. The detailed statistics of these datasets are summarized in Table~\ref{table_dataset}.

\begin{table*}[]
\begin{center}
\vspace{-4mm}
\renewcommand\arraystretch{1.5}
\caption{Running time in seconds of all algorithms on all temporal networks. $\delta = 600$s and $\#threads=1$. }
\vspace{-1mm}
\label{tab.single_thread}
\setlength{\tabcolsep}{1.6mm}{}

\begin{tabular}{c|c|c|cc|c|c|cc|c|cc}
\toprule
\multirow{2}{*}{Dataset} & EX    & EWS  & \multicolumn{2}{c|}{FAST}           & BT-Pair    & BTS-Pair   & \multicolumn{2}{c|}{FAST-Pair}            & 2SCENT-Tri & \multicolumn{2}{c}{FAST-Tri}         \\ \cline{2-12} 
                         & Time (s)  & Time (s) & \multicolumn{1}{c|}{Time (s)} & speedup & Time (s) & Time (s) & \multicolumn{1}{c|}{Time (s)} & speedup & Time (s)     & \multicolumn{1}{c|}{Time (s)} & speedup \\ \midrule
Email-Eu                 & 0.4739   & 0.3685  & \multicolumn{1}{c|}{0.4324}  &  \multicolumn{1}{c|}{1.1x}   & 0.6903  & 0.1754  & \multicolumn{1}{c|}{0.0679}  &  \multicolumn{1}{c|}{10.1x} &   11.2839  & \multicolumn{1}{c|}{0.1885}  &  59.8x \\ \hline
CollegeMsg               & 0.0847   & 0.0621  & \multicolumn{1}{c|}{0.0560}  &    1.5x   & 0.1604  & 0.0312  & \multicolumn{1}{c|}{0.0186}  &   8.6x      &   1.4527  & \multicolumn{1}{c|}{0.0218}  & 66x     \\ \hline
Bitcoinotc               & 0.1147   & 0.0280  & \multicolumn{1}{c|}{0.0170}  & 6.5x    & 0.1108  & 0.0037  & \multicolumn{1}{c|}{0.0054}  & 20.5x    &    1.1021   & \multicolumn{1}{c|}{0.0074}  &    148x    \\ \hline
Bitcoinalpha             & 0.0703   & 0.0195  & \multicolumn{1}{c|}{0.0113}  & 6.2x    & 0.0772  & 0.0106  & \multicolumn{1}{c|}{0.0051}  & 15.1x    &   0.75487      & \multicolumn{1}{c|}{0.0046}  &  164x   \\ \hline
Act-mooc                 & 2.9573   & 0.6978  & \multicolumn{1}{c|}{1.4810}  & 1.9x    & 0.6823  & 0.3587  & \multicolumn{1}{c|}{0.2032}  &   3.4x     &   3.5363  & \multicolumn{1}{c|}{0.8942}  &    3.9x     \\ \hline
SMS-A                    & 0.4060   & 0.5008  & \multicolumn{1}{c|}{0.1956}  & 2.1x    & 0.8308  & 0.1262  & \multicolumn{1}{c|}{0.0925}  & 8.9x    &   4.6176  & \multicolumn{1}{c|}{0.0351} &    131x     \\ \hline
FBWALL                   & 1.2045   & 0.7607  & \multicolumn{1}{c|}{0.1339}  & 8.8x    & 1.0071  & 0.1173  & \multicolumn{1}{c|}{0.0583}  & 17.3x   &   6.2523   & \multicolumn{1}{c|}{0.0578}  &    108x     \\ \hline
MathOverflow             & 2.5498   & 0.3468  & \multicolumn{1}{c|}{0.1013}  & 25x     & 1.2335  & 0.0952  & \multicolumn{1}{c|}{0.0180}  & 68.5x   &  3.2846 & \multicolumn{1}{c|}{0.0426}  &   77x   \\ \hline
Askubuntu                & 3.0907   & 0.6341  & \multicolumn{1}{c|}{0.2681}  & 11x     & 2.4465  & 0.2393  & \multicolumn{1}{c|}{0.0562}  & 43.5x    &   5.9988  & \multicolumn{1}{c|}{0.1750}  &   34x   \\ \hline
Superuser                & 5.6000   & 0.9960  & \multicolumn{1}{c|}{0.4199}  & 13x     & 3.7189  & 0.5203  & \multicolumn{1}{c|}{0.0835}  & 26.7x    &   8.3369  &  \multicolumn{1}{c|}{0.2808}  &    29x    \\ \hline
WikiTalk                 & 27.1527  & 12.1908 & \multicolumn{1}{c|}{23.0929} & 1.2x    & 18.8316  & 7.5769  & \multicolumn{1}{c|}{4.5980}  & 4.1x    &    61.8645   & \multicolumn{1}{c|}{17.3112} & 3.5x     \\ \hline
IA-online-ads            & 80.5552  & 48.9832 & \multicolumn{1}{c|}{26.3587} & 2.6x    & 28.4565 & 13.3298 & \multicolumn{1}{c|}{7.1345}   & 3.9x    & 183.3022   & \multicolumn{1}{c|}{2.1798} & 84.1x \\ \hline
StackOverflow            & 503.9162 & 57.2630 & \multicolumn{1}{c|}{59.5124} & 8x      & 97.3559 & 26.8346 & \multicolumn{1}{c|}{4.9405}   & 19.7x   & 215.6287   & \multicolumn{1}{c|}{49.9042} & 4.3x    \\ \hline
Rec-MovieLens            & 1238.8557 & 149.399 & \multicolumn{1}{c|}{136.2383} & 9.1x  & 83.6547 & 24.9987 & \multicolumn{1}{c|}{20.6102} & 4x & 484.9321 & \multicolumn{1}{c|}{63.6283} & 7.6x \\ \hline
Soc-bitcoin              & 3491.3959 & 595.5708 & \multicolumn{1}{c|}{802.7460} & 4.3x  & 698.2412 & 86.4989 & \multicolumn{1}{c|}{98.8155}  & 7.1x  & 1795.2359 & \multicolumn{1}{c|}{198.9950} & 9x \\  \hline
RedditComments          & 7968.3687 & 1133.7942 & \multicolumn{1}{c|}{1019.3465} & 7.8x  & 1605.1336 & 123.0630 & \multicolumn{1}{c|}{158.9597}  & 10.1x  & 2943.4058 & \multicolumn{1}{c|}{360.0926} & 8.2x \\ \bottomrule
\end{tabular}
\end{center}
\vspace{-4mm}
\end{table*}

\vspace{-1mm}
\subsection{Baselines}
We compare our algorithms against the following baselines:
\begin{itemize}
\item \textbf{EX}~\cite{paranjape2017motifs} -- An exact algorithm for counting all 2-node and 3-node, 3 edge temporal motifs. This algorithm is most relevant to our problem and is our main competitor in some experiments. 

\item \textbf{2SCENT}~\cite{kumar20182scent} -- An algorithm for enumerating simple temporal cycle, \ie, triangle temporal motifs. 

\item \textbf{BT}~\cite{mackey2018chronological} -- A backtracking algorithm for temporal subgraph isomorphism. 
Since there is no public code, we use the BT algorithm implemented by~\cite{liu2019sampling} for counting pair temporal motifs. 

\item \textbf{BTS}~\cite{liu2019sampling} -- An approximate algorithm based on interval sampling for temporal motif counting with exact algorithm \textbf{BT} used as a subroutine. 

\item \textbf{EWS}~\cite{wang2020efficient} -- An approximate algorithm based on edge and wedge sampling for counting temporal motifs with 3 nodes and 3 edges. 

\end{itemize}

Notice that we use *-Pair and *-Tri to denote the variants of * for counting pair temporal motifs and triangle temporal motifs, respectively.

\vspace{-1mm}
\subsection{Experimental Setting}
\label{sec.setting}

All experiments are conducted on a server running Ubuntu 18.10 with 40-core 2.30GHz Intel Xeon E5-2650 v3 processor and 128GB RAM. 
We download the codes of baselines published by the authors and followed the compilation and usage instructions. 
All algorithms are implemented in C++11 compiled by GCC v8.3.0 with -O3 optimizations. 
To run in parallel, all baselines use the same OpenMP mode to our method for parallel speedup.  

For 2SCENT, we use the algorithm with bloom filter and bundle to find the temporal cycles. 
For BTS, we use the same parameter sampling probability $q_j$ provided in~\cite{liu2019sampling}. 
For EWS, we set the edge sampling $p=0.01$ and wedge sampling $q=1$. 

\subsection{Accuracy Evaluation} 

First, we verify the accuracy of the proposed \model on four datasets with different degree distributions and different data scales compared with the exact EX algorithm. 
We report the counts of motif instances of all 2-node and 3-node, 3-edge temporal motifs with $\delta=600s$ detected by our method and EX algorithm in Fig.~\ref{fig.instances}. 


As we can see, the color scale in blue figures and red figures are identical, which means our \model can find out the same number of motif instances of all kinds of temporal motifs as exact EX algorithm on four tested datasets.  
For example, both \model and EX count 302K motif instances of all star temporal motifs on the small dataset CollegeMsg, and both detect 1217K motif instances of all triangle temporal motifs on the large dataset StackOverflow in all.  
In addition, our \model can detect accurate number of instances for specific type of temporal motif, \eg, pair temporal motifs. Our \model detect 31.6K instances and 10.6K instances of pair motif $M_{55}$ on CollegeMsg and Superuser datasets, which are the same as those of EX.  
Furthermore, our \model can detect accurate results on extremely unbalanced dataset (\ie, WikiTalk). For example, \model find out exact 65.7M and 64.9M instances for star temporal motifs $M_{13}$ and $M_{53}$ on WikiTalk compared to EX algorithm. 
Lastly, this experiment also verifies that our designed triple and quadruple counters for both star/pair and triangle motifs all can accurately identify the kinds of different temporal motifs.

\begin{figure*}[h]
    \centering
    \vspace{-4mm}
    \hspace{-6mm}
    \includegraphics[width=1.03\linewidth]{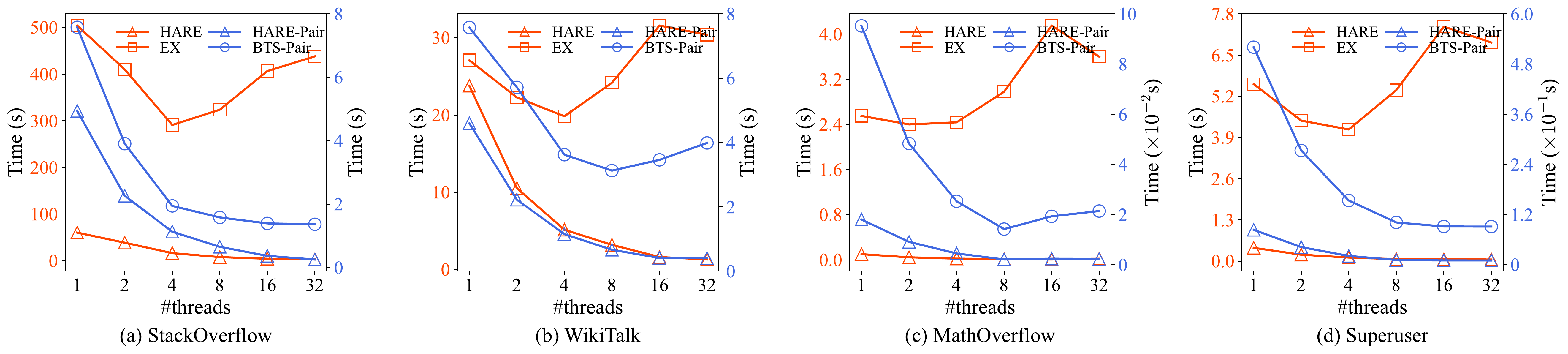}\\
    \vspace{-5mm}
    \hspace{-6mm}    
    \includegraphics[width=1.03\linewidth]{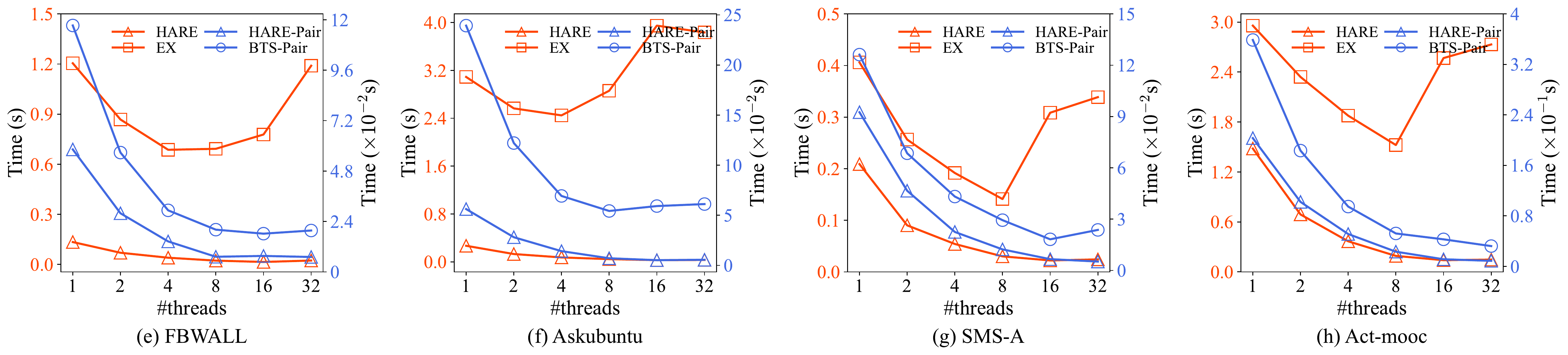}\\
    \vspace{-5mm}
    \hspace{-6mm}
    \includegraphics[width=1.03\linewidth]{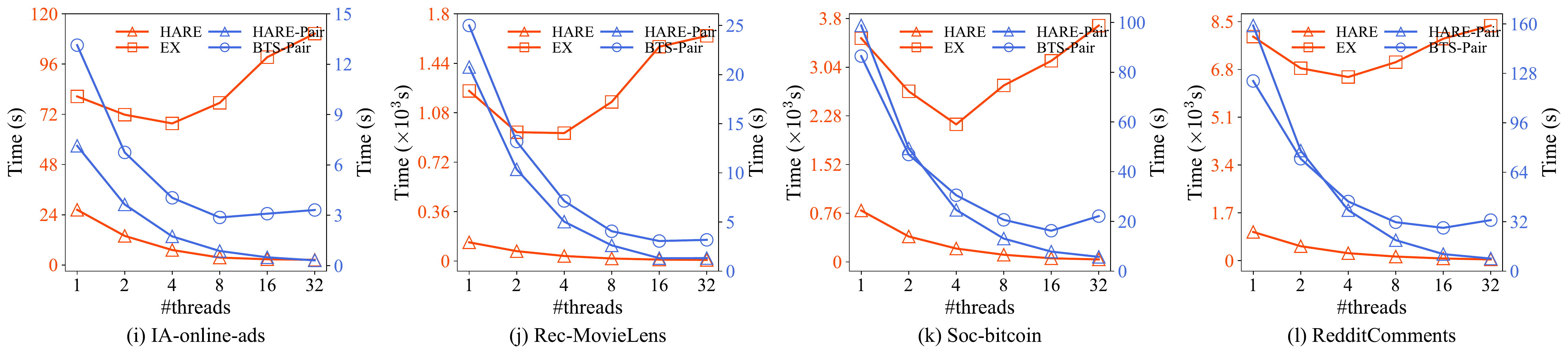}\\
    \vspace{-1mm}
    \caption{Running time in seconds of parallel algorithms \wrt. $\#threads$.}
    \label{fig.Parallel}
    \vspace{-2mm}
\end{figure*}

\subsection{Efficiency Evaluation} 
\label{sec.efficiency}

We next study the efficiency of our \model in counting temporal motifs on all datasets compared with both exact and sampling baselines in a single-threaded environment. We report the experimental results of all algorithms in Table~\ref{tab.single_thread}. Notice that EX, BT-Pair, and 2SCENT-Tri are exact algorithms, and EWS and BTS-Pair are sampling algorithms. Since our \model is an exact algorithm, the speedup is calculated by the corresponding exact algorithm. 2SCENT can only detect the triangle motif $M_{26}$, while \modelt can find out all kinds of triangle temporal motifs.

As shown in Table~\ref{tab.single_thread}, our \model (including \modelp and \modelt) is significantly faster than all exact baselines on all datasets in the single-threaded environment. In particular, \model achieves average $7\times$ speedup against EX algorithm across all datasets, reaching up to $25\times$ speedup on MathOverflow. 
This is because \model can directly identify the kinds of temporal motifs according to edge information and relationship between edges, and uses two quadruple counters (\ie, $Star[\cdot,\cdot,\cdot,\cdot]$ and $Tri[\cdot,\cdot,\cdot,\cdot]$) and one triple counter (\ie, $Pair[\cdot,\cdot,\cdot]$) to record the number of instances for all kinds of temporal motifs simultaneously. 
Besides, our \model also employs the most efficient implementation strategies to achieves the least edge traversal and counter updating operation, which significantly reduces the computation cost.   
Although EX algorithm also achieves linear time complexity, it maintains more than ten triple and tuple counters and requires multiple complex update operations for each temporal edge.  
Our \modelp and \modelt also perform much better than their corresponding exact algorithms (\ie, BT-Pair and 2SCENT-Tri). Specifically, \modelp and \modelt improve average $18.2\times$ and $65.7\times$ efficiency against their counterparts on all datasets, even if 2SCENT-Tri only count one kind of triangle temporal motif. \modelp achieves more than 100 speedup over 2SCENT-Tri on four datasets (\ie, Bitcoinotc, Bitcoinalpha, SMS-A and FBWALL). The main reason is that 2SCENT-Tri detects the triangle motif $M_{26}$ by enumerating all circles, and thus the time complexity is very large. Our \modelt detects all triangle temporal motifs in the edge sequence of each node recursively, which can achieve a running time linear in the number of temporal edges in the graph.  


Moreover, our \model achieves the comparable performance to state-of-the-art sampling algorithms (\ie, EWS and BTS). On ten of fourteen datasets (\eg, SMS-A, Facebook-wall, MathOverflow, Askubuntu, Superuser and Rec-MovieLens), our \model even significantly exceeds state-of-the-art sampling algorithm EWS. 
For pair motif counting, our \modelp also significantly outperforms the sampling algorithm BTS-Pair on almost all datasets, reaching $6.2\times$ speedup on MathOverflow dataset. 
The possible reason is that our propose algorithm not only designs the ingenious counters, but also achieves the least edge traversal without any redundancy in single threading, so as to minimize computation cost. 
Due to the large time complexity of the basic algorithms adopted by the sampling methods, the time consumption of the sampling methods is still relatively large. For example, on large-scale StackOverflow and Rec-MovieLens datasets, the sampling methods no longer has any time efficiency advantage in comparison to our \model.


\subsection{Scalable Evaluation}
\label{sec.scale}





Next, we evaluate the scalability of our proposed hierarchical parallel framework \framework compared with the parallel baselines. We report the experimental results on twelve datasets in Fig.~\ref{fig.Parallel}. Notice that \framework and EX algorithm refer to left ordinate axis, and \frameworkp and BTS-Pair refer to right ordinate axis. 
For our hierarchical parallel framework, we set $thr_{d}$ to the minimum value of degrees of top 20 nodes in each dataset. 


As shown in Fig.~\ref{fig.Parallel}, our \framework consistently outperforms EX algorithm in all tested cases on all tested datasets. 
As the number of threads increases, the counting time of \framework decreases, almost reaching a linear speedup. In particular, \framework achieves $26.3\times$ and $24\times$ speedup on large-scale StackOverflow and RedditComments datasets as the number of threads increases from 1 to 32. 
But the time consumed by Ex first decreases and then increases as the number of threads increases. After more than 16 threads, the running time of EX even exceeds the time of single thread on most datasets. 
This is because our designed \model realizes the divide and conquer of the nodes in the network, and solves the problem of data dependence, which is very suitable for parallel computing. At the same time, the proposed hierarchical parallel framework \framework achieves simultaneous inter-node and intra-node parallelism, which effectively solves the problem of unbalanced load caused by nodes with a large degree consuming most of the time.
Although EX algorithm can be implemented in parallel by dividing different time periods, its algorithm design involves data dependence, such as the counting processing for each edge depends on the situations of the previous edges. According to this design, each thread cannot be completely independent. When more threads are used, the time consumption caused by data dependency would offset the improvement brought by multi-threading. 
Specifically, our \framework achieves $538\times$, $203\times$, $148\times$, and $137\times$ speedup against EX algorithm when $\#threads = 32$ on MathOverflow, RedditComments, Rec-MovieLens, and SuperUser datasets, respectively.

Furthermore, our \frameworkp also significantly performs better than the sampling method BTS-Pair for counting pair temporal motifs in multi-threaded environment. 
As the number of threads increases, the time utilized by both \frameworkp and BTS-Pair decreases. However, the time consumed by BTS-Pair begins to increase on some datasets (\eg, FBWALL and WikiTalk), while our \frameworkp is decreasing. 
In particular, \frameworkp is faster than BTS-Pair by 10.6 times on large-scale IA-online-ads dataset when $\#threads = 32$. 
The main reason may be that BT algorithm employed by BTS has large time complexity (\ie, $O(|\mathcal{E}| (d^{\delta})^{|\mathcal{E}|-1})$), while our algorithm for star/pair motifs is linear in the number of temporal edges in the graph, \ie, $O(2 d^{\delta}|\mathcal{E}|)$).

In summary, this experiment effectively verifies the scalability of our proposed hierarchical framework on multi-threaded parallelism. Meanwhile, it also illustrates the huge advantages and potential capabilities of our framework on large-scale network datasets.

\subsection{Parameter Sensitivity}
\label{sec.parameter}

\begin{figure}[h]
    \centering
    \vspace{-3mm}
    \hspace{-5mm}
    \subfigure[Time \wrt. $\delta$]{
    \includegraphics[width=0.505\linewidth]{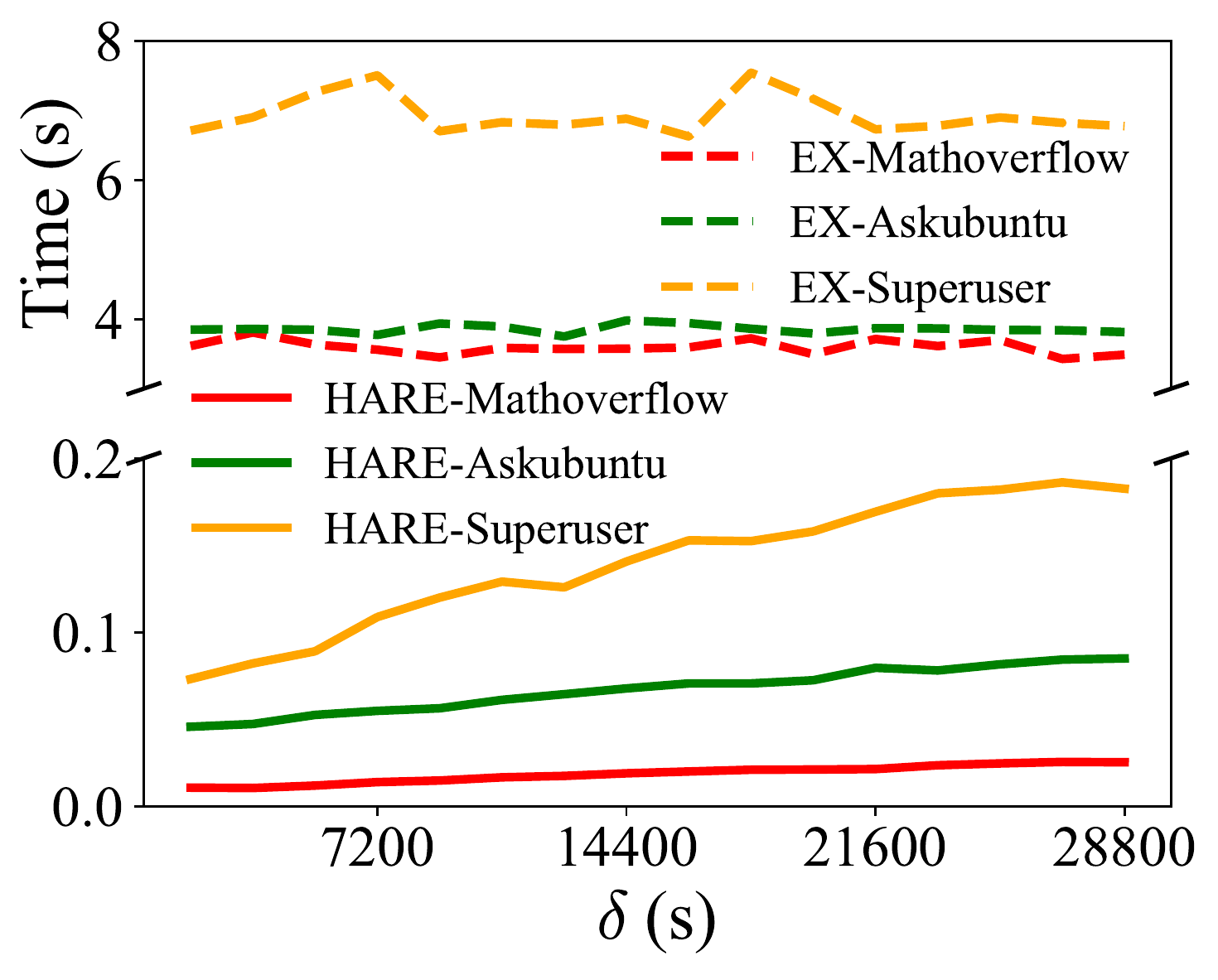}
    }
    \hspace{-5mm}
    \subfigure[Time \wrt. $thr_d$]{
    \includegraphics[width=0.505\linewidth]{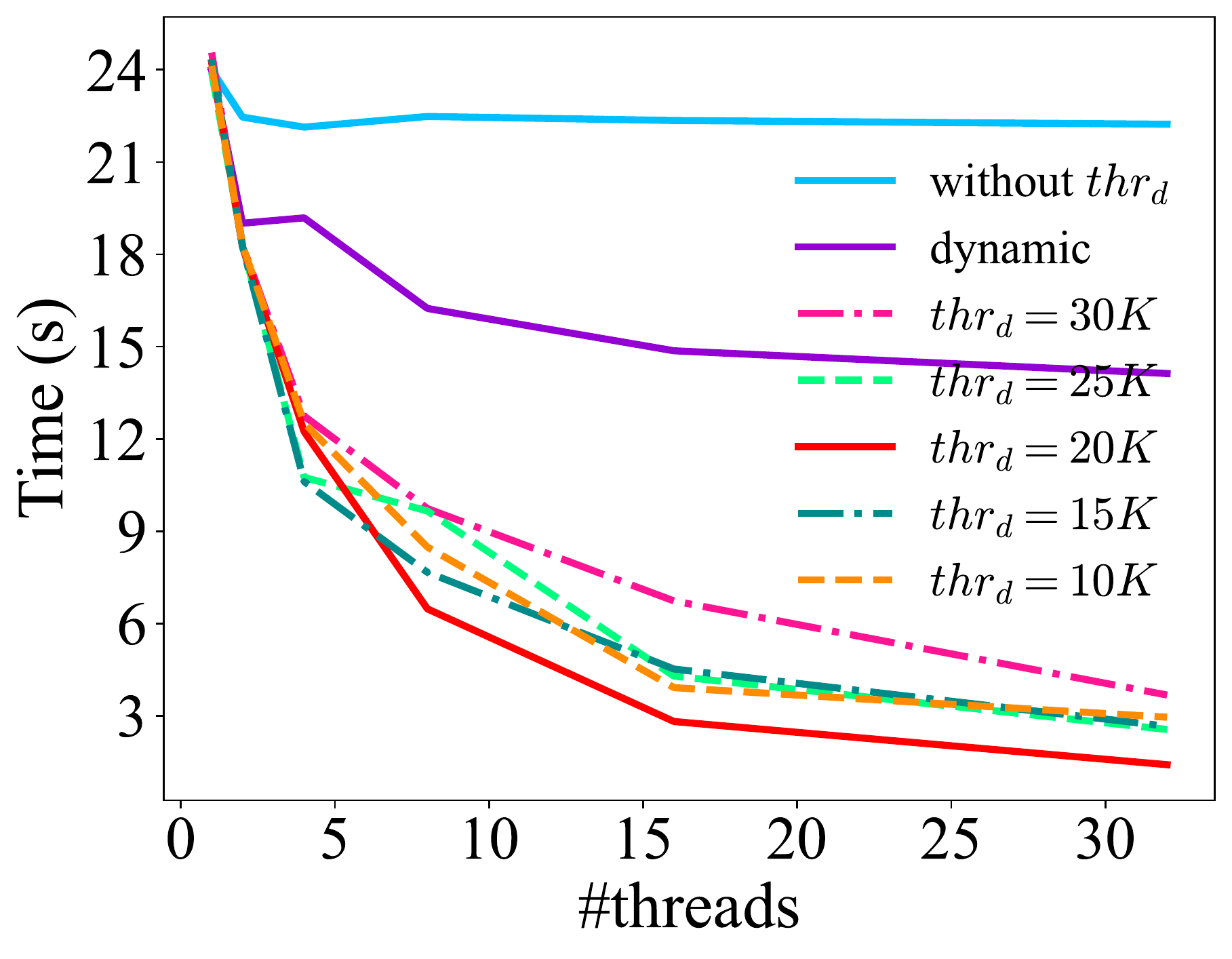}
    }
    \vspace{-2mm}
    \caption{Parameter sensitivity \wrt. $\delta$ and $thr_d$}
    \label{fig.sensitivity}
    \vspace{-2mm}
\end{figure}

We now investigate the sensitivity of our proposed framework \framework \wrt. two important parameters, \ie, time constraint $\delta$ and the degree threshold $thr_d$. 
We report the running time of \framework and EX with different $\delta$ values on SuperUser, Askubuntu, and MathOverflow datasets when $\#threads = 32$ in Fig.~\ref{fig.sensitivity}(a), and report the running time of \framework \wrt. different $thr_d$ varying $\#threads$ from 1 to 32 on WikiTalk dataset in Fig.~\ref{fig.sensitivity}(b), where `dynamic' denotes that dynamic scheduling model of OpenMP is used, and `without $thr_d$' means that default (\ie, static) mode is used.

From the results in Fig.~\ref{fig.sensitivity}(a), we can see that the performance of EX is almost unaffected by $\delta$, and our \framework increases slightly with the increase of $\delta$. This is consistent with the above time complexity analysis. 
However, our \framework is still 37-138 times faster than EX algorithm on three datasets when $\delta = 28.8K$ seconds. 
Fig.~\ref{fig.sensitivity}(b) illustrates the performance of our parallel framework \wrt. the values of degree threshold $thr_d$. 
As you can see, our \framework that uses the hierarchical parallel strategy is significantly faster than the parallel variations without it. Specifically, \framework reaches the best performance when $thr_d=20K$ among all tested cases. When $thr_d=20K$ and $thr_d=25K$, our algorithm performance has some decline, but still performs good. That is, our algorithm can obtain an expected speedup when $thr_d$ falls in a large range. 
Additionally, we also verify the effectiveness of dynamic scheduling model of OpenMP on WikiTalk dataset. As shown in Fig.~\ref{fig.sensitivity}(b), dynamic scheduling indeed significantly outperforms the version without it (\ie, dynamic vs. without $thr_d$).

\section{Conclusion}

In this paper, we propose a scalable solution for temporal motif counting in large-scale temporal networks. 
Specifically, we design two fast exact algorithms \models and \modelt for counting motif instances for star/pair temporal motifs and triangle temporal motifs, respectively. 
Our customized \models and \modelt both achieve the time complexity linear in the number of temporal edges of input graph. 
Based on the natural parallelism of our designed two fast algorithms, we eventually propose a hierarchical parallel framework \framework that fully leverages the multi-threading capacity of modern CPU to realize the most efficient temporal motif counting. 
We perform extensive experiments on sixteen real-world temporal graphs to demonstrate the superiority and scalability of our proposed method compared with other baselines. Our proposed framework \framework achieves up to $538$ times faster than state-of-the-art techniques. 
Although the proposed algorithms are specially optimized for 2- and 3-node, 3-edge temporal motifs, it will be able to efficiently count the higher-order (more nodes) temporal motifs by expanding the number of center nodes and slightly adapting the structure of the counters, which will be our future work. 


\section*{Acknowledgment}
This work is partially supported by the National Natural Science Foundation of China under grant Nos. 62176243, 61773331 and 41927805, the National Program on Key Research Project  under grant No. 2019YFC1509100, and the National Key Research and Development Program of China under grant No. 2018AAA0100602. 

%
\IEEEpeerreviewmaketitle



%
\clearpage
\balance
\bibliographystyle{IEEEtran}
\bibliography{icde_reference}


\end{document}